\pdfoutput=1
\documentclass[11pt]{article}

\usepackage[preprint]{acl}
\usepackage[utf8]{inputenc}
\DeclareUnicodeCharacter{66F9}{}
\DeclareUnicodeCharacter{5E7F}{}
\DeclareUnicodeCharacter{7855}{}

\usepackage{times}
\usepackage{latexsym}

\usepackage[T1]{fontenc}

\usepackage[utf8]{inputenc}

\usepackage{microtype}
\usepackage{inconsolata}
\usepackage{graphicx}

\PassOptionsToPackage{hyphens}{url}

\usepackage{booktabs,longtable,array,makecell}
\renewcommand{\arraystretch}{1.15}

\newcolumntype{L}[1]{>{\raggedright\arraybackslash}p{#1}}
\newcolumntype{C}[1]{>{\centering\arraybackslash}p{#1}}

\usepackage{caption}
\usepackage{xcolor}
\usepackage{tikz}
\usetikzlibrary{positioning,arrows.meta}
\definecolor{famBlue}{HTML}{2F71B5}
\definecolor{famPurple}{HTML}{7A52C7}
\definecolor{famTeal}{HTML}{2A9D8F}
\definecolor{famOrange}{HTML}{E07A2B}
\definecolor{famRed}{HTML}{D1495B}
\definecolor{subGray}{HTML}{F2F2F2}

\newcommand{\wFamily}{0.17\textwidth}
\newcommand{\wSub}{0.22\textwidth}
\newcommand{\wModels}{0.30\textwidth}

\usepackage{titlesec}

\titlespacing*{\section}
  {0pt}{0.8ex plus .2ex minus .2ex}{0.6ex plus .2ex}   

\titlespacing*{\subsection}
  {0pt}{0.6ex plus .2ex minus .2ex}{0.4ex plus .2ex}

\titlespacing*{\subsubsection}
  {0pt}{0.5ex plus .2ex minus .2ex}{0.3ex plus .1ex}

\usepackage{enumitem}
\setlist{nosep,leftmargin=1.5em}

\title{LLM4Cell: A Survey of Large Language and Agentic Models for Single-Cell Biology}

\author{
\textbf{Sajib Acharjee Dip}\thanks{\,Equal contribution.}\textsuperscript{1},
\textbf{Adrika Zafor}\footnotemark[1]\textsuperscript{2},
\textbf{Bikash Kumar Paul}\textsuperscript{1}\\[4pt]
\textbf{Uddip Acharjee Shuvo}\textsuperscript{3},
\textbf{Muhit Islam Emon}\textsuperscript{1},
\textbf{Xuan Wang}\textsuperscript{1},
\textbf{Liqing Zhang}\thanks{\,Correspondence: \texttt{sajibacharjeedip@vt.edu}, \texttt{lqzhang@cs.vt.edu}.}\textsuperscript{1,4}\\[5pt]
\textsuperscript{1}Department of Computer Science, Virginia Tech, Blacksburg, VA, USA\\
\textsuperscript{2}Department of Computational Modeling and Data Analytics, Virginia Tech, Blacksburg, VA, USA\\
\textsuperscript{3}Institute of Information and Technology, University of Dhaka, Dhaka, Bangladesh\\
\textsuperscript{4}Fralin Biomedical Research Institute at VTC: Cancer Research Center, Washington DC, USA
}

\begin{document}
\maketitle
\begin{abstract}
Large language models (LLMs) and emerging agentic frameworks are beginning to transform single-cell biology by enabling natural-language reasoning, generative annotation, and multimodal data integration.  
However, progress remains fragmented across data modalities, architectures, and evaluation standards.  
\textbf{LLM4Cell} presents the first unified survey of 58 foundation and agentic models developed for single-cell research, spanning RNA, ATAC, multi-omic, and spatial modalities.  
We categorize these methods into five families—foundation, text-bridge, spatial/multimodal, epigenomic, and agentic—and map them to eight key analytical tasks including annotation, trajectory and perturbation modeling, and drug-response prediction.  
Drawing on over 40 public datasets, we analyze benchmark suitability, data diversity, and ethical or scalability constraints, and evaluate models across 10 domain dimensions covering biological grounding, multi-omics alignment, fairness, privacy, and explainability.  
By linking datasets, models, and evaluation domains, LLM4Cell provides the first integrated view of language-driven single-cell intelligence and outlines open challenges in interpretability, standardization, and trustworthy model development.
\end{abstract}

\section{Introduction}

Large language models (LLMs) are transforming biomedical discovery by linking molecular patterns with knowledge encoded in text.  ~\citep{zhang2025survey, aljohani2025comprehensive, bi2024ai}. Prior work on computational workflows and biological prediction tasks demonstrates how learned representations capture complex molecular relationships and enhance practical performance. ~\citep{lam2024large, bhattacharya2024large, dip2024deepage, nam2024using, shuvo2024assessing} 
In single-cell biology, where each experiment measures thousands of genes across millions of cells, LLMs promise to unify gene expression, chromatin accessibility, and spatial organization under a shared, interpretable framework.  
Early foundation models such as \textbf{scGPT} ~\citep{cui2024scgpt}, \textbf{Geneformer} ~\citep{theodoris2023transfer}, and \textbf{scFoundation} ~\citep{hao2024large} learn transferable representations directly from single-cell data, while emerging systems like \textbf{CellLM} ~\citep{zhao2023large}, \textbf{scAgent} ~\citep{mao2025scagent}, and \textbf{CellVerse} ~\citep{zhang2025cellverse} extend these capabilities toward reasoning, dialogue, and autonomous analysis.  
Together, they signal a shift from statistical inference to language-driven, interpretable, and generalizable cell intelligence.

Despite this progress, the field remains fragmented.  
Models differ widely in data modality, supervision type, and evaluation standard ranging from scRNA-seq to ATAC, multi-omic, and spatial measurements hindering cross-model comparison and reproducibility.  
Benchmarking is inconsistent, datasets are unevenly distributed across modalities, and agentic frameworks lack standardized ways to measure reasoning correctness or biological grounding.  
Without a systematic synthesis, it is difficult to assess what current single-cell LLMs truly understand or where they fall short.

To address these challenges, we introduce \textbf{LLM4Cell}, a comprehensive survey of large language and agentic models for single-cell biology.  
We analyze 58 representative methods spanning five methodological families\textbf{Foundation}, \textbf{Text-Bridge}, \textbf{Spatial/Multimodal}, \textbf{Epigenomic}, and \textbf{Agentic} and organize them across eight major tasks, from annotation and trajectory inference to perturbation and drug-response modeling.  
Complementing this, we compile over 40 publicly available datasets across RNA, ATAC, multi-omic, spatial, perturbation, and plant domains, and evaluate each model using a ten-dimension rubric covering grounding, fairness, scalability, and interpretability.

\paragraph{Contributions.}
\textbf{LLM4Cell} provides:
(1) a modality-balanced registry of around 40 benchmark datasets for large-model training and evaluation;  
(2) a unified taxonomy of 58 foundation and agentic models across eight analytical tasks;  
(3) a ten-dimension domain rubric capturing biological grounding, generalization, and ethics; and  
(4) a critical discussion of open challenges in cross-modal alignment, reasoning, and trustworthy AI for cell biology.

\section{Related Work}

The intersection of large language models (LLMs) and single-cell biology has gained rapid momentum, motivating several early surveys and benchmarks.  
The ACL~2025 survey on foundation models for single-cell biology~\citep{zhang2025survey} outlines pretrained and fine-tuned architectures across protein, gene, and cell representations.  
Broader reviews such as~\citet{lan2025large} and~\citet{bian2024general} discuss ``large cellular models'' (LCMs) and trends in scaling, tokenization, and transfer learning across biological data.  
Benchmark studies including \textit{Single-Cell Omics Arena}~\citep{liu2024single} and \textit{CellVerse}~\citep{zhang2025cellverse} evaluate LLMs on annotation and question answering, while works like \textit{scInterpreter}~\citep{guo2023scinterpreter} and \textit{scExtract}~\citep{wu2025scextract} demonstrate practical integration into annotation and curation pipelines.

However, existing reviews remain limited in scope.  
They focus on architectures or prompting but give little attention to the \textit{datasets} underpinning model training and evaluation especially those covering spatial, perturbation, or multimodal modalities.  
Standardized benchmarks, domain-aware metrics, and reproducible splits are seldom addressed, and plant single-cell datasets key for cross-kingdom generalization are almost entirely absent.  
Moreover, agentic reasoning and tool-augmented systems are typically treated as isolated examples rather than a coherent methodological class, leaving the relationship among data modalities, modeling paradigms, and biological domains fragmented.

\textbf{LLM4Cell} advances this literature by providing an integrated, data-centric synthesis linking models, datasets, and evaluation domains.  
Unlike prior descriptive taxonomies, it unifies over forty public datasets and fifty-eight models across modalities and tasks, assessing their biological grounding, fairness, and scalability.  
In doing so, \textbf{LLM4Cell} establishes a reproducible foundation for cross-modal benchmarking and future research on trustworthy, autonomous single-cell intelligence.

\section{Datasets}

Progress in large language models for single-cell biology relies on the rapid growth of high-quality datasets across transcriptomic, epigenomic, multimodal, and spatial domains.  
Our survey compiles over forty public resources spanning five major modalities \textbf{RNA}, \textbf{ATAC}, \textbf{multi-omic}, \textbf{spatial}, and \textbf{perturbation/drug-response} plus emerging \textbf{plant single-cell atlases}.  
These datasets underpin model pretraining, evaluation, and cross-modal reasoning for annotation, integration, trajectory inference, perturbation modeling, and spatial mapping.

\textbf{Transcriptomic atlases} such as \textit{Tabula Sapiens}, \textit{Tabula Muris} ~\citep{tabula_sapiens2022}, and the \textit{Human Cell Atlas} ~\citep{travaglini2020_hlca} remain dominant for foundation-model training and ontology-aware annotation.  
\textbf{Chromatin-accessibility data} (e.g.\ \textit{Cusanovich mouse atlas}~\citep{cusanovich2018_science}, human \textit{adult/fetal scATAC}~\citep{zhang2021_encode}) map regulatory states but remain sparse and heterogeneous.  
\textbf{Multi-omic resources} (e.g.\ \textit{TEA-seq} ~\citep{swanson2021_natbiotech}, \textit{DOGMA-seq} ~\citep{mimitou2021scalable}, \textit{CITE-seq}~\citep{stoeckius2017simultaneous}) link RNA, ATAC, and protein modalities, providing supervision for cross-view alignment.  
\textbf{Spatial technologies} (e.g.\ \textit{Visium} ~\citep{oliveira2025high}, \textit{Slide-seqV2} ~\citep{stickels2021highly}, \textit{MERFISH} ~\citep{chen2015spatially}, \textit{Stereo-seq} ~\citep{chen2022spatiotemporal}) connect molecular profiles to tissue architecture, while \textbf{functional datasets} (e.g.\ \textit{Perturb-seq}~\citep{replogle2022_cell}) benchmark causal reasoning.  
\textbf{Plant resources} (e.g.\ \textit{scPlantDB} ~\cite{he2024sty}, \textit{Arabidopsis E-CURD-4} ~\citep{shulse2019high}) extend modeling beyond animal systems, opening cross-kingdom evaluation.

Despite this diversity, major gaps persist.  
\textit{(i)} RNA atlases vastly outscale other modalities.  
\textit{(ii)} Benchmark fragmentation and inconsistent metadata hinder reproducibility.  
\textit{(iii)} Privacy constraints limit access to clinical spatial datasets.  
\textit{(iv)} Non-human and plant data remain scarce.  
\textit{(v)} Paired and tri-modal resources provide ideal testbeds for next-generation multimodal and agentic LLMs. Table \ref{tab:datasets-rna}, \ref{tab:datasets-atac}, \ref{tab:datasets-multiome}, 
\ref{tab:datasets-spatial},
\ref{tab:datasets-perturb}, \ref{tab:datasets-plant} (\textit{Appendix D}) list representative datasets with tasks, scale, and source links for reproducible benchmarking.

\section{Model Taxonomy}

\begin{figure*}[t]
\centering
\scriptsize
\begin{tikzpicture}[
  box/.style={draw,rounded corners,align=left,inner xsep=6pt,inner ysep=5pt},
  fam/.style   ={box,minimum height=11mm,text width=\wFamily,font=\bfseries\scriptsize,text=white},
  sub/.style   ={box,fill=subGray,draw=black!30,text width=\wSub},
  mdl/.style   ={box,fill=white,draw=black!20,text width=\wModels},
  arr/.style   ={->,>=Stealth,very thin}
]

\newcommand{\TaxRow}[9]{%
  \begin{scope}[yshift=#1]
    \node[fam,fill=#2] (F) {#3};

    \node[sub] (S1) [right=10mm of F] {#4};
    \node[sub] (S2) [below=5mm of S1,xshift=0mm] {#6};
    \node[sub] (S3) [below=5mm of S2,xshift=0mm] {#8};

    \node[mdl] (M1) [right=8mm of S1] {#5};
    \node[mdl] (M2) [right=8mm of S2] {#7};
    \node[mdl] (M3) [right=8mm of S3] {#9};

    \draw[arr] (F.east) -- (S1.west);
    \draw[arr] (F.east) |- (S2.west);
    \draw[arr] (F.east) |- (S3.west);
    \draw[arr] (S1.east) -- (M1.west);
    \draw[arr] (S2.east) -- (M2.west);
    \draw[arr] (S3.east) -- (M3.west);
  \end{scope}
}

\TaxRow{0mm}{famBlue}{Foundation Models}
{Self-supervised (masked / next-token)}
{\textit{scGPT}, \textit{Geneformer}, \textit{scFoundation}, \textit{scBERT}, \textit{tGPT}, \textit{scTrans}, \textit{scRDiT}, \textit{scGraphformer}, \textit{scGFT}}
{Multi-task / adapters}
{\textit{CellFM}, \textit{scPRINT}, \textit{scMulan}, \textit{scMoFormer}, \textit{scmFormer}}
{Cross-species}
{\textit{iSEEEK}, \textit{UCE}, \textit{GeneCompass}}

\TaxRow{-28mm}{famPurple}{Text-Bridge \& Ontology LLMs}
{Cell-to-text}
{\textit{Cell2Sentence (C2S)}, \textit{C2S-Scale}, \textit{Cell2Text}}
{Text-conditioned embeddings}
{\textit{GenePT (GENEPT)}, \textit{scELMo}, \textit{scInterpreter}}
{Ontology / prompt alignment}
{\textit{CellLM}, \textit{QuST-LLM}}

\TaxRow{-56mm}{famTeal}{Spatial \& Multimodal FMs}
{RNA + Spatial}
{\textit{TransformerST}, \textit{stFormer}, \textit{scGPT-spatial}, \textit{Spatial2Sentence}, \textit{spaCCC}}
{RNA + ATAC + Protein}
{\textit{scMMGPT}, \textit{OmiCLIP}, \textit{FmH2ST}}
{Histology-conditioned}
{\textit{spaLLM}, \textit{HEIST}, \textit{Nicheformer}}

\TaxRow{-84mm}{famOrange}{Epigenomic \& Regulatory FMs}
{Chromatin accessibility}
{\textit{EpiFoundation}, \textit{EpiBERT}, \textit{EpiAttend}}
{Efficient / long-range}
{\textit{GeneMamba}, \textit{scMamba}}
{Cross-omic regulatory}
{\textit{ChromFound}, \textit{GET}}

\TaxRow{-112mm}{famRed}{Agentic \& Reasoning Frameworks}
{LLMs as annotation agents}
{\textit{scAgent}, \textit{LICT}}
{Reasoning + tools}
{\textit{EpiAgent}, \textit{ChatCell}, \textit{scExtract}}
{Interactive / multi-agent}
{\textit{CellVerse}, \textit{Teddy}, \textit{Pilot}}

\end{tikzpicture}
\caption{
Hierarchical taxonomy for \textit{LLM4Cell}.
Color-coded families expand into sub-branches and representative models, tracing the progression from foundation pretraining to multimodal and agentic reasoning frameworks. References are omitted for visibility and included in the appendix method comparison table.
}
\label{fig:taxonomy}
\end{figure*}
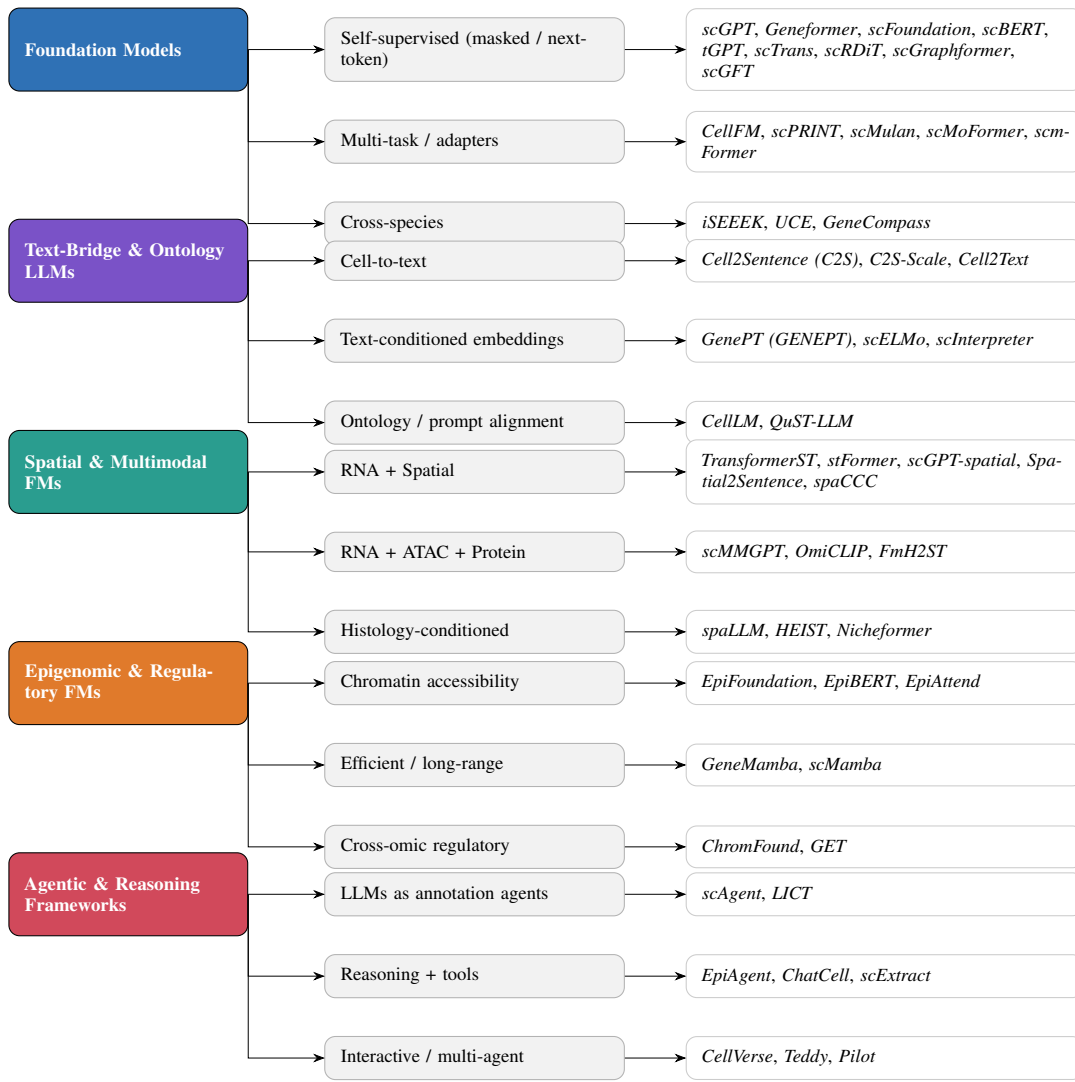

Large language models are rapidly reshaping single-cell biology, producing diverse architectures, pretraining schemes, and reasoning frameworks.  
We categorize 58 representative methods into five methodological families \textbf{Foundation Models}, \textbf{Text-Bridge LLMs}, \textbf{Spatial and Multimodal Models}, \textbf{Epigenomic Models}, and \textbf{Agentic Frameworks} based on their core design and data modality.  
These span the progression from gene-level embeddings to multimodal and autonomous systems capable of biological reasoning.

Our taxonomy is organized along five orthogonal dimensions: (i) \textit{Modality} (RNA, ATAC, multi-omic, spatial, or hybrid), (ii) \textit{Grounding type} (atlas, ontology, or marker-based), (iii) \textit{Agentic capability} (multi-step reasoning or autonomous orchestration), (iv) \textit{Primary task} (annotation, trajectory, perturbation, integration, etc.), and (v) \textit{Domain quality}, represented by ten quantitative scores for grounding, fairness, scalability, and interpretability.

Figure~\ref{fig:taxonomy} summarizes this taxonomy, with detailed attributes in Appendix~Table~\ref{tab:methods-full}.  
Foundation models currently dominate in scope and adoption, while emerging spatial, epigenomic, and agentic frameworks signal a shift toward contextual, reasoning-driven single-cell intelligence.

\subsection{Foundation Models}
Foundation models learn transferable cell and gene embeddings directly from large-scale scRNA-seq without explicit labels.  
Representative systems such as \textbf{scGPT} ~\citep{cui2024scgpt}, \textbf{Geneformer} ~\citep{theodoris2023transfer}, and \textbf{scFoundation} ~\citep{hao2024large} pretrain on multi-tissue atlases exceeding one million cells, using masked-gene or rank-based reconstruction to capture expression context.  
Variants like \textbf{tGPT} ~\citep{shen2023generative}, \textbf{scBERT} ~\citep{yang2022scbert}, and \textbf{scGraphformer} ~\citep{fan2024scgraphformer} treat genes as tokens or graph nodes, while cross-species models (\textbf{UCE} ~\citep{rosen2023universal}, \textbf{GeneCompass}~\citep{yang2024genecompass}) apply contrastive alignment across homologous genes.  
These models underpin most single-cell LLM pipelines, offering strong transfer for annotation and integration but limited ontology grounding and explainability, motivating later text-bridged and reasoning frameworks.

\begin{figure*}[t]
    \centering
    \includegraphics[width=\textwidth]{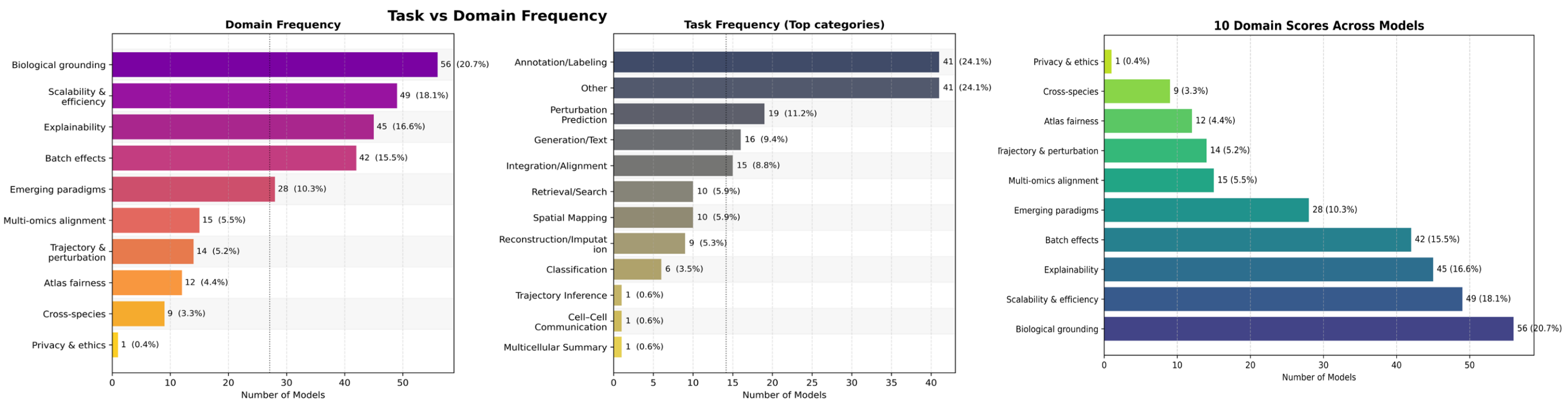}
    \caption{
        Task and domain distribution across spatial transcriptomics models. Most emphasize biological grounding, scalability, and annotation tasks.
    }
    \label{fig:all_ablation}
\end{figure*}

\subsection{Text-Bridge LLMs}
Text-Bridge models couple molecular embeddings with biomedical language to ground single-cell representations in semantics and ontology.  
Systems such as \textbf{scELMo} ~\citep{liu2023scelmo}, \textbf{CellLM} ~\citep{zhao2023large}, and \textbf{GenePT} ~\citep{chen2024genept} align gene or cell embeddings with textual descriptors, while \textbf{Cell2Sentence} ~\citep{pmlr-v235-levine24a} and \textbf{Cell2Text} ~\citep{kharouiche2025cell2text} translate expression profiles into natural-language summaries.  
Using dual-encoder or encoder–decoder architectures, these frameworks employ contrastive or prompt-based alignment between molecular and text encoders (e.g., BioBERT).  
They enhance interpretability and enable zero-shot annotation but depend heavily on curated corpora and remain non-agentic, serving as a bridge between foundation and reasoning frameworks.

\subsection{Spatial and Multimodal Models}
Spatial and multimodal frameworks integrate gene expression with spatial coordinates, histology, or additional omics to capture tissue architecture.  
Representative models include \textbf{TransformerST} ~\citep{lu2024sctrans}, \textbf{spaLLM}~\citep{10778152}, and \textbf{OmiCLIP}~\citep{cui2025towards} for spatial mapping, and \textbf{scMMGPT}~\citep{shi2025multimodal} and \textbf{FmH2ST} for RNA–ATAC–protein integration.  
They use multi-branch Transformers with modality-specific encoders and cross-attention fusion, trained on datasets such as Visium DLPFC and MERFISH.  
These models achieve strong spatial alignment and biological realism but face heterogeneous resolutions, limited open benchmarks, and high computational cost.

\subsection{Epigenomic Models}
Epigenomic foundation models extend LLM concepts to chromatin-accessibility and regulatory data such as scATAC-seq.  
\textbf{EpiFoundation} ~\citep{wu2025epifoundation}, \textbf{EpiBERT}~\citep{javed2025multi}, and \textbf{EpiAttend}~\citep{li2022epiattend} learn cis-regulatory patterns from ENCODE and other compendia, while \textbf{GeneMamba} and \textbf{scMamba} use efficient state-space layers for long-range dependency modeling.  
Cross-omic variants (\textbf{ChromFound}~\citep{jiao2025chromfound}, \textbf{GET}~\citep{fu2025foundation}) jointly embed RNA and ATAC to infer gene-regulatory networks.  
These models improve biological grounding but remain constrained by sparse data and lack unified benchmarks across regulatory modalities.

\subsection{Agentic Frameworks}
Agentic systems integrate pretrained models with reasoning modules for autonomous single-cell analysis.  
Frameworks such as \textbf{scAgent}~\citep{mao2025scagent}, \textbf{CellVerse}~\citep{zhang2025cellverse}, and \textbf{Teddy}~\citep{chevalier2025teddy} combine domain-specific encoders with LLM controllers that plan tasks, query ontologies, and interface with tools or APIs.  
\textbf{EpiAgent}~\citep{chen2025epiagent} extends this paradigm to regulatory genomics.  
These systems enable dialogue-based annotation and multi-step reasoning but lack standardized benchmarks for reasoning fidelity and rely on underlying LLM reliability.

\section{Task-specific Applications}

While the previous section categorized models by architectural family, here we analyze them through the lens of \textit{biological objectives} shown in Figure \ref{fig:all_ablation} and Appendix Figure \ref{fig:heatmap}.  
Across the 58 methods surveyed, we identify eight recurring tasks that define single-cell modeling pipelines: (1) annotation and ontology mapping, (2) trajectory and perturbation modeling, (3) multi-omic integration, (4) spatial mapping and deconvolution, (5) regulatory-network and pathway inference, (6) cross-species translation, (7) generative simulation, and (8) drug-response prediction.  
Most models span multiple tasks-for example, \textit{scGPT} and \textit{Geneformer} support both annotation and perturbation prediction-reflecting the convergence between representation learning and functional reasoning.  

This task-centric view complements the architectural taxonomy by revealing how foundation and agentic systems differ in operational scope.  
Foundation models dominate annotation, integration, and trajectory inference, whereas emerging text-bridge and agentic systems extend to knowledge-grounded reasoning and dynamic planning.  
Spatial and epigenomic frameworks contribute specialized capabilities in tissue mapping and chromatin-level interpretation.

\subsection{Annotation and Ontology Mapping}

Annotation is the most common single-cell task, spanning automated cell-type labeling, ontology alignment, and cross-dataset harmonization.  
Foundation models such as \textbf{scGPT}, \textbf{scFoundation}, and \textbf{scBERT} achieve high accuracy on atlases like Tabula~Sapiens and the Human~Cell~Atlas.  
Text-bridge models (\textbf{CellLM} ~\citep{zhao2023large}, \textbf{scELMo} ~\citep{liu2023scelmo}, \textbf{GenePT} ~\citep{chen2024genept}) add ontology-based interpretability, and agentic systems (\textbf{scAgent}) perform multi-step annotation reasoning via LLM controllers.  
Models are fine-tuned on curated references (e.g., Azimuth~PBMC, HCA~Lung) using masked-gene or contrastive objectives.  
Evaluation uses label accuracy or ARI; key limitations include rare-cell detection, cross-species consistency, and lack of reasoning benchmarks.

\subsection{Trajectory and Perturbation Modeling}

Trajectory modeling captures dynamic state transitions and causal responses to interventions.  
\textbf{Geneformer} ~\citep{theodoris2023transfer}, \textbf{scGPT}~\citep{cui2024scgpt}, and \textbf{scRDiT} ~\citep{dong2025scrdit} model temporal or perturbation trajectories using denoising and contrastive objectives, trained on datasets such as Replogle~2022 Perturb-seq and sci-Plex.  
Epigenomic models (\textbf{EpiFoundation}~\citep{wu2025epifoundation} ,\textbf{EpiAgent}~\citep{chen2025epiagent}) extend to regulatory responses through chromatin context and ontology reasoning.  
Metrics include expression-profile correlation and target-gene recovery.  
While these models predict single perturbations effectively, performance degrades for combinatorial or long-range effects, underscoring the need for unified causal benchmarks.

\subsection{Multi-omic Integration}

Multi-omic integration seeks unified representations across paired RNA, ATAC, and protein modalities to capture gene-regulatory and signaling relationships.  
Representative models include \textbf{scMMGPT}~\citep{shi2025multimodal}, \textbf{GET (General Expression Transformer)~\citep{fu2025foundation}}, \textbf{ChromFound}~\citep{jiao2025chromfound}, and epigenomic-aware variants such as \textbf{EpiFoundation}~\citep{wu2025epifoundation} and \textbf{EpiBERT}~\citep{javed2025multi}.  
These systems train on joint datasets like 10x Multiome PBMC 10k, TEA-seq, DOGMA-seq, and ASAP-seq, using cross-modal transformers or contrastive alignment between expression and accessibility features.

Most architectures employ modality-specific encoders with shared latent spaces or token-type embeddings; \textit{scMGPT} applies masked-reconstruction across modalities, while \textit{GET} learns cross-attention between gene and chromatin tokens.  
Compared with unimodal models, they show improved cell-type alignment and batch correction but face data-scale imbalance and sparse modality overlap.  
Evaluation commonly uses modality-matching accuracy and latent-space correlation.  
Despite strong integration performance, interpretability and standard benchmarks remain limited, and scalability to tri-modal data is an active challenge.

\subsection{Spatial Mapping and Deconvolution}

Spatial mapping links molecular profiles to their tissue locations, enabling inference of cell organization and microenvironmental context.  
Representative models include \textbf{TransformerST} ~\citep{lu2024sctrans}, \textbf{spaLLM} , \textbf{OmiCLIP}, \textbf{FmH2ST}, \textbf{HEIST} ~\citep{madhu2025heist}, and \textbf{Spatial2Sentence}~\citep{chen2025spatial}, trained on datasets such as Visium DLPFC, Slide-seqV2, MERFISH, and Xenium.  
They combine spatial coordinates, histology features, and gene-expression embeddings through cross-attention or contrastive alignment to reconstruct cell or spot-level maps.

\textit{TransformerST} performs axial attention between gene and spatial axes for spot-to-cell deconvolution, while \textit{spaLLM} and \textit{OmiCLIP} align histology and transcriptomic features using language-guided contrastive learning.  
Models like \textit{HEIST} and \textit{FmH2ST} fuse H\&E images with RNA profiles for tissue-domain segmentation.  
Evaluation typically measures spatial correlation, clustering accuracy, or cell-type F1 against curated annotations.  
Although these models improve contextual understanding of tissue architecture, limited spatial benchmarks and heterogeneous resolutions constrain systematic comparison across methods.

\subsection{Regulatory Network and Pathway Inference}

Regulatory-network and pathway inference focuses on identifying gene–gene or enhancer–promoter dependencies that drive transcriptional programs.  
Representative models include \textbf{GeneMamba}~\citep{qi2025bidirectional}, \textbf{scMamba}~\citep{yuan2025scmamba}, \textbf{EpiFoundation}~\citep{wu2025epifoundation}, \textbf{EpiBERT}~\citep{javed2025multi}, \textbf{ChromFound}~\citep{jiao2025chromfound}, and \textbf{GET}~\citep{fu2025foundation}, trained on large-scale ATAC and multiome atlases such as ENCODE, human cCRE datasets, and 10x Multiome PBMC.  
These models treat chromatin regions or genes as tokens and learn context-dependent regulatory relationships through self- or cross-attention.

\textit{GeneMamba}~\citep{qi2025bidirectional} and \textit{scMamba} employ state-space layers for efficient modeling of long-range chromatin dependencies, while \textit{EpiBERT} and \textit{EpiFoundation} use masked-region reconstruction to capture enhancer–promoter coupling.  
\textit{ChromFound} and \textit{GET} link RNA and accessibility signals to infer transcription-factor activity and causal directionality.  
Evaluation typically uses motif enrichment, AUROC for known interactions, or pathway-level overlap with KEGG or Reactome references.  
These models improve biological grounding and causal interpretability but remain constrained by sparse training data and lack of unified ground-truth regulatory benchmarks.

\subsection{Cross-Species Translation}

Cross-species translation aims to transfer cell and gene representations between organisms, enabling comparative and evolutionary analysis of single-cell atlases.
Representative models include \textbf{iSEEEK} ~\citep{shen2022universal}, \textbf{UCE} (Universal Cell Embeddings) ~\citep{rosen2023universal}, \textbf{GeneCompass} ~\citep{yang2024genecompass}, and \textbf{scPlantLLM} ~\citep{scplantlmm}, trained on paired or homolog-mapped datasets such as tthe Mouse Cell Atlas, Tabula Muris, and scPlantDB. These systems learn alignment between orthologous genes or conserved cell states to support zero-shot annotation across species.

\textit{iSEEEK} ~\citep{shen2022universal} integrates over 11 million human and mouse cells using shared gene-token vocabularies, while \textit{UCE} ~\citep{shen2023generative} applies masked autoencoding and contrastive alignment across ortholog groups.  
\textit{GeneCompass} ~\citep{yang2024genecompass} extends this with knowledge-informed pretraining that links species via ontology-derived embeddings, and \textit{scPlantLLM} adapts the paradigm to plant single-cell data across 17 species.  
Evaluation uses transfer accuracy and embedding consistency across species pairs.  
Despite improved cross-domain generalization, these models remain limited by incomplete ortholog mappings and domain shifts in sequencing depth and tissue composition.

\subsection{Generation and Simulation}

Generative modeling aims to synthesize realistic single-cell profiles, simulate perturbations, or reconstruct missing modalities.  
Representative approaches include \textbf{scGPT}, \textbf{scFoundation}, \textbf{CellFM} ~\citep{zeng2025cellfm}, and \textbf{Geneformer}, which employ decoder or diffusion-style architectures to model the probability distribution of gene-expression states.  
These models are pretrained on large atlases such as Tabula~Sapiens and HCA, then fine-tuned to generate new cell embeddings or perturbation outcomes.

\textit{scGPT} and \textit{scFoundation} use masked-token prediction extended with generative decoding to simulate unseen cells or conditions, while \textit{CellFM} introduces multimodal conditioning for cross-tissue synthesis.  
Generative evaluation typically measures distributional similarity (KL divergence, FID-like embedding distance) or recovery of known cell-type proportions.  
Such models enable scalable in-silico experiments and data augmentation but face challenges in controlling biological realism and avoiding overfitting to dominant cell types.

\subsection{Drug-Response Prediction}

Drug-response prediction models aim to infer cellular transcriptional outcomes following chemical or genetic perturbations, supporting virtual screening and mechanism discovery.  
Representative frameworks include \textbf{Geneformer}~\citep{theodoris2023transfer}, \textbf{scGPT} ~\citep{cui2024scgpt}, \textbf{EpiFoundation}~\citep{wu2025epifoundation}, and \textbf{EpiAgent}~\citep{chen2025epiagent}, trained or evaluated on large-scale pharmacotranscriptomic resources such as sci-Plex, Replogle 2022 Perturb-seq, and the ARC Virtual Cell Challenge.  
These systems learn conditional embeddings that map control to perturbed states, enabling zero-shot prediction of unseen compounds or targets.

\textit{Geneformer} predicts post-treatment expression profiles using masked-token denoising, while \textit{scGPT} and \textit{EpiFoundation} integrate modality-specific conditioning to capture epigenetic or pathway context.  
\textit{EpiAgent} further incorporates ontology-guided reasoning to link molecular features with literature-derived drug–gene relationships.  
Evaluation relies on Pearson correlation or top-$k$ recovery of known targets between predicted and observed responses.  
Although these models demonstrate strong predictive capacity for single-agent perturbations, performance degrades for combinatorial treatments and unseen pathways, highlighting the need for richer multimodal pharmacogenomic benchmarks.

\noindent\textbf{Cross-task observations.}
Across all eight task domains, clear convergence patterns emerge.  
Foundation models dominate annotation, integration, and generative simulation owing to large-scale pretraining on transcriptomic atlases, whereas text-bridge systems extend interpretability through ontology grounding and natural-language supervision.  
Spatial and epigenomic models expand biological realism by capturing tissue architecture and regulatory logic, while agentic frameworks introduce multi-step reasoning and tool orchestration that connect these modalities into autonomous workflows.  
Despite methodological diversity, evaluation remains fragmented-annotation and integration enjoy standardized benchmarks, but trajectory, drug-response, and agentic reasoning lack common protocols.  
Cross-species and privacy-aware tasks are the least represented, reflecting data scarcity and regulatory constraints.  
Overall, task analysis reveals a rapidly maturing ecosystem in which foundation and multimodal pretraining supply general representations, and emerging agentic paradigms begin to operationalize biological reasoning.  
The following section quantifies these trends through ten domain-level dimensions encompassing biological grounding, fairness, scalability, and interpretability.

\section{Domain Analysis}

We evaluate fifty-eight methods across ten domain dimensions capturing reliability, generalization, and interpretability.  
Each score reflects published evidence of performance or explicit design alignment, summarized in Appendix~Table~\ref{tab:methods-full} and Figure \ref{fig:all_ablation}.  
Together, these dimensions provide a composite view of model maturity beyond task accuracy.

\subsection{Biological Grounding}
Most foundation and multimodal models achieve strong biological grounding by encoding marker genes, pathways, or regulatory regions.  
\textbf{scGPT}, \textbf{Geneformer}, and \textbf{EpiFoundation} leverage atlas-scale training to recover known cell types and pathway enrichments.  
Spatial frameworks such as \textbf{TransformerST} and \textbf{OmiCLIP} add tissue-context grounding via histology alignment.  
However, textual and ontology grounding remain inconsistent, and biological semantics are rarely explicit in model objectives.

\subsection{Batch Effects and Heterogeneity}
Robustness to batch effects varies widely.  
\textbf{scFoundation} ~\citep{hao2024large}, \textbf{CellFM} ~\citep{zeng2025cellfm}, and \textbf{iSEEEK} ~\citep{shen2022universal} mitigate donor and protocol variation through large-scale multi-domain pretraining, while others rely on explicit batch tokens or adversarial alignment.  
True cross-platform generalization remains limited; even large models show degradation when tested on unseen sequencing chemistries or tissue types.

\subsection{Multi-omics Alignment}
Models such as \textbf{scMGPT} ~\citep{palayew2025towards}, \textbf{GET} ~\citep{fu2025foundation}, and \textbf{ChromFound} ~\citep{jiao2025chromfound} demonstrate consistent alignment across RNA, ATAC, and protein features using cross-attention or contrastive fusion.  
Spatial hybrids (\textbf{HEIST}, \textbf{FmH2ST}) further connect molecular and visual domains.  
While alignment quality improves, limited paired datasets and uneven modality coverage constrain reproducibility and benchmarking.

\subsection{Trajectory and Perturbation}
Dynamic inference is strongest in generative transformers (\textbf{Geneformer}, \textbf{scGPT}) and epigenomic variants (\textbf{EpiAgent}).  
They predict post-perturbation expression or accessibility shifts with reasonable fidelity but struggle to model multi-target or time-continuous responses.  
Few methods quantify causal uncertainty or biological plausibility.

\subsection{Cross-Species and Cross-Tissue Generalization}
Cross-organism transfer remains challenging.  
\textbf{iSEEEK} ~\citep{shen2022universal}, \textbf{UCE} ~\citep{rosen2023universal}, and \textbf{GeneCompass} ~\citep{yang2024genecompass} align human and mouse cell embeddings via ortholog mapping, while \textbf{scPlantLLM} ~\citep{scplantlmm} extends this to plants.  
Performance drops sharply for divergent taxa, reflecting incomplete ortholog tables and bias toward human datasets.

\subsection{Atlas Fairness and Representation Balance}
Dataset imbalance affects nearly all models.  
Human and immune-cell–dominant atlases overrepresent certain tissues and demographics.  
\textbf{Teddy}~\citep{chevalier2025teddy} introduces benchmarking for fairness and representation balance, but fairness metrics are rarely adopted.  
No model yet enforces demographic or tissue-specific parity during training.

\subsection{Explainability and Interpretability}
Text-bridge systems (\textbf{GenePT}, \textbf{CellLM}, \textbf{Cell2Text} ~\citep{kharouiche2025cell2text}) improve interpretability by linking embeddings to ontology terms.  
Agentic systems (\textbf{scAgent}) extend this via natural-language reasoning and tool explanations.  
Most foundation models still rely on attention visualization, providing limited causal transparency.

\subsection{Privacy and Ethics}
Data privacy remains underexplored.  
Synthetic generation (\textbf{scGFT} ~\citep{nouri2025single}) and open-access chat systems (\textbf{ChatCell} ~\citep{fang2024chatcell}) raise concerns over re-identification and data leakage.  
No single-cell LLM currently implements federated or privacy-preserving learning, and ethical guidance for multi-omic sharing remains informal.

\subsection{Scalability and Efficiency}
Scaling trends mirror NLP: efficient transformers (\textbf{xTrimoGene}~\citep{gong2023xtrimogene}) and state-space models (\textbf{GeneMamba}~\citep{qi2025bidirectional}, \textbf{scMamba}~\citep{yuan2025scmamba}) reduce memory cost while retaining accuracy.  
\textbf{CellFM} ~\citep{zeng2025cellfm} demonstrates training across 100 M cells, showing feasibility of atlas-scale learning.  
Still, high compute requirements limit accessibility for most research groups.

\subsection{Emerging Paradigms and Agentic Behavior}
Recent frameworks (\textbf{scAgent} ~\citep{mao2025scagent}, \textbf{CellVerse} ~\citep{zhang2025cellverse}, \textbf{EpiAgent}~\citep{chen2025epiagent}) introduce reasoning and autonomous decision pipelines, integrating LLM controllers with specialized encoders.  
These systems achieve the highest scores in explainability and cross-modal planning but lack standardized evaluation of reasoning fidelity or reproducibility.

\section{Open Problems}

\subsection{Trust and Validation}
Evaluation across modalities remains inconsistent.  
Metrics often emphasize reconstruction over biological plausibility, and few studies offer independent replication.  
Community-curated leaderboards and standardized validation sets are essential for credible comparison.

\subsection{Data and Bias}
Training corpora are dominated by human and mouse atlases, limiting cross-species and clinical generalization.  
Rare-cell, plant, and microbial systems remain underrepresented, reinforcing demographic and biological bias.  
Diverse, balanced datasets are critical for equitable model development.

\subsection{Cross-Modal and Dynamic Modeling}
True integration of RNA, ATAC, spatial, and temporal modalities is still elusive.  
Existing models handle only pairwise combinations, while four-way fusion remains computationally infeasible.  
Unified tokenization and multimodal pretraining pipelines could enable consistent cross-domain reasoning.

\subsection{Interpretability and Causality}
LLM embeddings capture statistical correlation but rarely mechanistic insight.  
Bridging attention to regulatory or causal networks requires explicit reasoning layers and experimental grounding.  
Combining symbolic or causal discovery modules with transformers is a key next step.

\subsection{Ethics and Privacy}
Open single-cell data sharing raises privacy and dual-use concerns.  
Few models include audit trails or consent-aware governance.  
Adapting federated and differential-privacy methods to biological data is urgently needed.

\subsection{Agentic and Interactive Systems}
Agentic frameworks can plan and reason but lack reliable benchmarks.  
Tasks such as multi-step reasoning, dataset retrieval, and model orchestration remain fragile.  
Robust metrics for reasoning accuracy, safety, and reproducibility will be crucial for trustworthy biological AI.

\section{Conclusions}
\textbf{LLM4Cell} presents the first unified survey of large language and agentic models for single-cell biology, linking datasets, architectures, and evaluation domains within a common framework.  
By analyzing 58 models and over 40 public datasets across RNA, ATAC, multi-omic, spatial, and perturbation modalities, we show how foundation-scale pretraining, multimodal grounding, and agentic reasoning are reshaping single-cell analysis.  
Our taxonomy and evaluation rubric reveal a transition from purely statistical modeling toward language-driven, interpretable, and increasingly autonomous systems that connect molecular data with biological knowledge.  
We hope this synthesis provides a reproducible reference for benchmarking, model selection, and the design of next-generation cellular foundation and reasoning models.

\section*{Limitations}

Despite its breadth, this study has several limitations.  
First, reported performance metrics vary widely across publications, preventing consistent quantitative comparison.  
Second, our ten-dimension rubric captures qualitative trends but not standardized numerical rankings.  
Third, access restrictions and licensing prevented inclusion of certain clinical or proprietary spatial datasets, while non-animal single-cell resources remain scarce.  
Fourth, model efficiency, compute cost, and hyperparameter sensitivity were not systematically analyzed.  
Finally, as multimodal and reasoning frameworks evolve rapidly, \textbf{LLM4Cell} reflects a snapshot in time rather than a static benchmark.  
Future work should establish community leaderboards, causal interpretability tests, and reasoning benchmarks to advance trustworthy and reproducible single-cell intelligence.

\section*{Acknowledgments}
This work was supported in part by Virginia Tech, the Department of Computer Science, and the U.S. National Science Foundation (NSF) under Awards \#2125798, \#2344169, and \#2319522. We gratefully acknowledge their support and resources that made this research possible.

\bibliography{main}

\clearpage
\appendix
\label{sec:appendix}

\section{Data Collection}

\subsection*{A.1 Search Strategy}
To construct a comprehensive registry of large language models (LLMs) and agentic frameworks in single-cell biology, we systematically screened both peer-reviewed and preprint literature from 2020–2025 across multiple sources.  
Searches were conducted on PubMed, Google Scholar, arXiv, and Semantic Scholar using combinations of domain- and method-specific keywords, including:

\begin{itemize}
    \item \textbf{General:} ``large language model'', ``foundation model'', ``LLM'', ``multimodal model'', ``generative AI''  
    \item \textbf{Domain-specific:} ``single-cell'', ``scRNA-seq'', ``scATAC'', ``multiome'', ``spatial transcriptomics'', ``perturb-seq'', ``cell atlas'', ``biomedical foundation model''  
    \item \textbf{Integration / reasoning:} ``ontology alignment'', ``text-bridge'', ``agentic framework'', ``biological reasoning'', ``cross-modal integration'', ``annotation model''  
\end{itemize}

After removing duplicates, the combined queries returned approximately \textbf{8,020 papers} from 2020–2025.  
Of these, 5,510 remained after filtering for English-language articles with available abstracts.  
We manually reviewed titles, abstracts, and model descriptions to identify works directly involving large-scale pretrained models, multimodal transformers, or language-based reasoning systems applied to single-cell or cellular-level omics data.

\subsection*{A.2 Inclusion and Screening Criteria}
Studies were included if they met at least one of the following criteria:
\begin{enumerate}
    \item Introduced or evaluated a large pretrained model ($\geq$10M parameters) for single-cell or multi-omic analysis.
    \item Employed textual grounding, ontology prompts, or natural-language interfaces to interpret single-cell data.
    \item Proposed agentic or reasoning-based frameworks for biological tasks (annotation, trajectory, perturbation, or integration).
    \item Released code, preprints, or benchmarks relevant to cellular foundation models.
\end{enumerate}

Methodological, biological, and dataset-based duplicates were merged under representative entries.  
This curation yielded \textbf{58 distinct models} across five methodological families: foundation, text-bridge, spatial/multimodal, epigenomic, and agentic frameworks.  
Each was annotated with task domain, modality, training data, and evaluation scope (Appendix Table~\ref{tab:methods-full}).

\subsection*{A.3 Inclusion of Preprints}
We included arXiv and bioRxiv preprints to ensure coverage of recent and influential models not yet published in peer-reviewed venues.  
In single-cell research, the majority of foundational architectures (e.g., \textit{scGPT}, \textit{scFoundation}, \textit{CellLM}) initially appeared as preprints months before journal acceptance.  
Given the field’s rapid evolution and small number of LLM-scale models, preprints represent essential early contributions to reproducibility and transparency.  
Each preprint was manually cross-verified for active GitHub or Zenodo links to confirm technical validity and data availability.

\subsection*{A.4 Resulting Corpus}
The final corpus spans 58 methods across 37 unique institutions, 40+ publicly accessible datasets, and 8 major analytical tasks.  
This dataset–method pairing forms the empirical foundation for the taxonomic and domain analyses presented in the main paper.
\section{Extended Dataset Summaries}

\subsection{RNA Atlases}
Foundational scRNA-seq atlases such as \textit{Tabula Sapiens} (>1 M cells, 24 tissues) and \textit{Tabula Muris Senis} (mouse, multi-organ across aging) provide references for annotation and batch-effect evaluation.  
Organ-specific datasets like the \textit{Human Lung Cell Atlas (HLCA)} and the \textit{Allen Brain Map} enable domain-specific benchmarking and ontology-aware evaluation.

\subsection{Chromatin and ATAC Data}
The \textit{Cusanovich mouse sci-ATAC} atlas, human adult/fetal scATAC atlases, and joint RNA–ATAC modalities (\textit{SHARE-seq}, \textit{SNARE-seq2}) characterize regulatory landscapes supporting trajectory and GRN inference, though sparsity limits large-scale pretraining.

\subsection{Multiome and Tri-Modal Data}
Datasets such as \textit{TEA-seq}, \textit{DOGMA-seq}, \textit{ASAP-seq}, and \textit{CITE-seq} pair transcriptomic, epigenomic, and proteomic modalities, enabling cross-modal translation and representation learning.  
The \textit{Multiome Benchmark Pack} aggregates >40 datasets for standardization but remains smaller than large RNA atlases.

\subsection{Spatial Transcriptomics and Imaging}
Platforms including \textit{10x Visium/HD}, \textit{Slide-seqV2}, and \textit{DBiT-seq} provide spatial gene-expression maps, while imaging technologies (\textit{MERFISH}, \textit{STARmap}, \textit{Stereo-seq}, \textit{CosMx}, \textit{Xenium}) capture near single-molecule resolution for tissue-level reasoning.

\subsection{Perturbation and Drug-Response Screens}
Large CRISPR-based datasets (\textit{Replogle 2022 Perturb-seq}, \textit{Norman 2019}, \textit{sci-Plex}) and community benchmarks (\textit{Virtual Cell Challenge}) quantify transcriptional responses for causal and policy evaluation in agentic frameworks.

\subsection{Plant Single-Cell Datasets}
\textit{scPlantDB} (67 datasets, 17 species, 2.5 M cells) and \textit{PlantscRNAdb} provide unified plant atlases; \textit{Arabidopsis E-CURD-4} and transgenic tobacco scRNA-seq datasets enable cross-kingdom and stress-response modeling.

\subsection{Critical Observations}
\textbf{(1)} RNA atlases dominate in scale;  
\textbf{(2)} Chromatin and spatial data are fragmented;  
\textbf{(3)} Privacy and licensing restrict clinical data;  
\textbf{(4)} Cross-species coverage is limited;  
\textbf{(5)} Paired modalities offer promising benchmarks for next-generation LLMs.

Representative datasets and links are provided in Tables A1–A3.
\section{Extended Details on Model Taxonomy}
\subsection{Foundation Models (Extended Details)}

Foundation-scale models constitute the base layer for language-driven single-cell analysis.  
They are trained on molecular profiles—typically scRNA-seq or integrated multi-organ datasets—to learn generalizable representations of cellular states, gene programs, and perturbations.

\paragraph{Training corpora and scale.}
Most use atlases containing $10^6$–$10^8$ cells.  
\textit{scGPT} and \textit{scFoundation} were pretrained on human–mouse atlases from Tabula~Sapiens and HCA; \textit{Geneformer} aggregates thousands of GEO datasets; \textit{iSEEEK} and \textit{CellFM} span 11–17 organs and species, enabling cross-tissue generalization.

\paragraph{Architectural variations.}
Most adopt Transformer backbones with gene tokens as contextual units.  
\textit{tGPT} and \textit{scBERT} encode ranked or discrete gene sequences, while \textit{Geneformer} and \textit{scFoundation} use masked-token modeling.  
Hybrid designs such as \textit{scGraphformer} ~\citep{fan2024scgraphformer} and \textit{scRDiT} ~\citep{dong2025scrdit} incorporate graph or diffusion-based attention, improving spatial and regulatory context.

\paragraph{Learning objectives.}
Common objectives include masked-gene prediction, expression denoising, rank-based reconstruction, and cross-species contrastive learning (\textit{UCE}, \textit{GeneCompass}).  
Optimization typically minimizes cosine or KL divergence across gene embeddings, similar to masked-language modeling in NLP.

\paragraph{Applications and limitations.}
Foundation embeddings transfer effectively to annotation, integration, and trajectory inference but lack explicit biological grounding.  
Ontology alignment and reasoning remain external, and interpretability is largely post-hoc (attention heatmaps, gene-ranking).  
These limitations prompted the development of Text-Bridge and Agentic frameworks discussed in later sections.

\subsection{Text-Bridge LLMs (Extended Details)}
These models explicitly connect molecular and textual modalities via ontology labels, pathway terms, or literature embeddings.  
\textit{scELMo} fuses ELMo-based gene metadata with cell latent vectors; \textit{CellLM} uses ontology-aware prompts for natural-language annotation; \textit{Cell2Sentence} and \textit{Cell2Text} map gene-expression vectors to descriptive sentences through cross-modal contrastive loss; and \textit{GenePT} jointly pre-trains on PubMed to embed genes and diseases in the same space.  
Architectures typically adopt Transformer encoders for molecular data and domain-tuned BERT variants for text, trained with contrastive or cosine-similarity objectives.  
These systems rank highest in \textit{Explainability} and \textit{Emerging paradigms} scores (Appendix Table \ref{tab:methods-full}) but remain limited by vocabulary coverage and the lack of full multi-omic grounding.

\subsection{Spatial and Multimodal Models (Extended Details)}
Spatial models combine molecular and positional information to learn tissue-aware embeddings.  
\textit{TransformerST} performs cross-scale attention between spatial spots and gene tokens; \textit{spaLLM} aligns histology-derived captions with expression vectors using LLM guidance; and \textit{OmiCLIP} links image and molecular embeddings via contrastive pre-training.  
Multi-omic extensions such as \textit{scMMGPT} and \textit{FmH2ST} integrate RNA, ATAC, and protein modalities within a shared latent space.  
Grounding typically relies on marker-based or atlas alignment rather than textual ontologies.  
Evaluation metrics include spot-level reconstruction accuracy and correlation with histological segmentation (DLPFC, Visium HD).  
These models show high \textit{biological grounding} but moderate \textit{explainability} due to visual-feature opacity.

\subsection{Epigenomic Models (Extended Details)}
Inputs include chromatin-accessibility matrices, motif sequences, and enhancer–promoter links.  
\textit{EpiFoundation} pre-trains on the ENCODE scATAC compendium; \textit{EpiBERT} encodes open-chromatin sequences using masked-region objectives; and \textit{EpiAttend} models enhancer–promoter coupling through cross-region attention.  
State-space architectures (\textit{scMamba}, \textit{GeneMamba}) enable efficient context propagation over tens of thousands of genomic peaks.  
\textit{ChromFound} and \textit{GET} jointly model RNA and ATAC modalities for regulatory inference.  
Grounding leverages atlas-derived cCRE annotations and transcription-factor motifs.  
Performance is strong for \textit{biological grounding} and \textit{multi-omics alignment}, but interpretability is limited to motif attention visualization.

\subsection{Agentic Frameworks (Extended Details)}
Agentic models couple (i) a pretrained molecular encoder, (ii) an LLM-based controller, and (iii) external tool interfaces.  
\textit{scAgent} executes multi-step annotation workflows via ontology queries; \textit{CellVerse} coordinates multiple domain agents for transcriptomic, spatial, and literature reasoning; \textit{EpiAgent} extends agentic control to enhancer–gene analysis; and \textit{Teddy}/\textit{scPilot}~\citep{joodaki2024detection} provide lightweight orchestrators for benchmarking pipelines.  
Grounding sources include the Human Cell Atlas ontology, UBERON, and PubMed.  
These systems score highest in \textit{Explainability} and \textit{Emerging paradigms}, marking a transition from static embeddings to interactive, goal-directed modeling.  
Open challenges include evaluation of reasoning accuracy, privacy, and reproducibility.

\onecolumn
\section{Datasets}
\begingroup

\setlength{\LTpre}{0pt}
\setlength{\LTpost}{0pt}

\begin{longtable}{@{}L{0.18\textwidth} L{0.16\textwidth} L{0.42\textwidth} L{0.12\textwidth} L{0.12\textwidth}@{}}
\caption{\textbf{RNAseq} single-cell datasets used in LLM-based single-cell research.}
\label{tab:datasets-rna}\\
\toprule
\textbf{Dataset} & \textbf{Tasks} & \textbf{Description} & \textbf{Scale} & \textbf{Link / Citation} \\
\midrule
\endfirsthead
\toprule
\textbf{Dataset} & \textbf{Tasks} & \textbf{Description} & \textbf{Scale} & \textbf{Link / Citation} \\
\midrule
\endhead
\midrule \multicolumn{5}{r}{\emph{Continued on next page}} \\
\endfoot
\bottomrule
\endlastfoot

Tabula Sapiens v2 &
Annotation, Integration &
Multi-organ \textit{human} atlas ($\sim$1.1\,M cells, 28 organs across 24 donors) capturing cell-type heterogeneity and cross-tissue transcriptional variation; droplet and plate-seq modalities. &
1.1\,M cells / 28 organs / 24 donors &
\href{https://tabula-sapiens.sf.czbiohub.org/}{CZ Biohub Portal} \\
\midrule

Tabula Muris (mouse, multi-organ) &
Annotation, Integration &
Mouse multi-organ atlas of gene expression combining droplet and FACS; enables cross-tissue comparison and batch integration. &
$\sim$100{,}000 cells / 20 organs &
\href{https://explore.data.humancellatlas.org/projects/e0009214-c0a0-4a7b-96e2-d6a83e966ce0}{HCA Project Page} \\
\midrule

Mouse Cell Atlas (scMCA) &
Annotation, Integration &
Comprehensive mouse single-cell atlas constructed with Microwell-seq; supports cell-type matching via scMCA tool. &
$>$400{,}000 cells / $>$40 tissues &
\href{https://bis.zju.edu.cn/MCA/atlas3.html}{MCA Portal (ZJU)} \\ \midrule

Human Lung Cell Atlas (HLCA v1.0) &
Annotation, Integration, Disease Modeling &
Integrated human lung reference built from 49 scRNA-seq datasets ($\sim$2.4\,M cells) across 16 studies; unified epithelial, immune, and stromal annotations. &
$\sim$2.4\,M / 49 datasets / 16 studies &
\href{https://explore.data.humancellatlas.org/projects}{HLCA v1.0 Portal}   \\ \midrule

HCA lung project (example project page) &
Annotation, Integration, Disease Modeling &
HCA lung cohort integrating $\sim$2.4\,M single cells from 49 datasets; harmonized annotations for airway, immune, and endothelial populations. &
2.4\,M cells / 49 datasets &
\href{https://explore.data.humancellatlas.org/projects/cc95ff89-2e68-4a08-a234-480eca21ce79}{HLCA Project Page} \\ \midrule

Allen Brain Map -- Cell Types RNA-seq (human \& mouse) &
Annotation, Cross-species, Trajectory &
Single-cell and single-nucleus transcriptomes from human and mouse brain regions; supports cross-species comparison and cell-type taxonomy benchmarking. &
$>$1.8\,M cells / multiple cortical \& subcortical regions &
\href{https://portal.brain-map.org/atlases-and-data/rnaseq}{Allen Brain Cell Types Database} 
\\ \midrule
Yale Lung Disease Cell Atlases (e.g., COPD) &
Annotation, Disease Modeling &
Single-cell RNA atlases of diseased human lungs (COPD, IPF) from Yale’s Kaminski lab, capturing altered epithelial, endothelial, and immune populations. &
$\sim$300k cells (IPF + COPD + controls) &
\href{https://explore.data.humancellatlas.org/projects/c16a754f-5da3-46ed-8c1e-6426af2ef625}{HCA/Yale Atlas} 

\end{longtable}
\endgroup

\vspace{1em}


\begingroup

\setlength{\LTpre}{0pt}
\setlength{\LTpost}{0pt}

\begin{longtable}{@{}L{0.18\textwidth} L{0.16\textwidth} L{0.42\textwidth} L{0.12\textwidth} L{0.12\textwidth}@{}}
\caption{\textbf{ATACseq} and chromatin-accessibility datasets used in LLM-based single-cell research.}
\label{tab:datasets-atac}\\
\toprule
\textbf{Dataset} & \textbf{Tasks} & \textbf{Description} & \textbf{Scale} & \textbf{Link / Citation} \\
\midrule
\endfirsthead
\toprule
\textbf{Dataset} & \textbf{Tasks} & \textbf{Description} & \textbf{Scale} & \textbf{Link / Citation} \\
\midrule
\endhead
\midrule \multicolumn{5}{r}{\emph{Continued on next page}} \\
\endfoot
\bottomrule
\endlastfoot

Mouse sci-ATAC-seq atlas (Cusanovich et al., 2018) &
Annotation, Trajectory, GRN inference &
Landmark single-cell ATAC-seq atlas profiling $\sim$100,000 nuclei across 13 adult mouse tissues using combinatorial indexing (sci-ATAC-seq); enables cross-tissue regulatory and lineage analysis. &
$\sim$100k nuclei / 13 tissues &
\href{https://www.ncbi.nlm.nih.gov/geo/query/acc.cgi?acc=GSE111586}{GSE111586, Science 2018} \\
\midrule

Human adult scATAC atlas (Zhang et al., 2021) &
Annotation, GRN, Cross-tissue integration &
Comprehensive single-cell chromatin accessibility atlas of adult human tissues, defining candidate cis-regulatory elements (cCREs) across 25 tissues and 222 cell types; foundation of the ENCODE human cCRE registry. &
$\sim$472k nuclei / 25 tissues &
\href{https://www.encodeproject.org/single-cell/}{Nature 2021, ENCODE Portal} \\ \midrule

Human fetal scATAC atlas (Domcke et al., 2020; GSE149683) &
Annotation, Developmental trajectory, GRN &
Single-cell ATAC-seq atlas of human fetal tissues profiling 15 organs across mid-gestation; reveals developmental enhancer activity and regulatory lineage trajectories. &
$\sim$720k nuclei / 15 organs &
\href{https://www.ncbi.nlm.nih.gov/geo/query/acc.cgi?acc=GSE149683}{Science 2020, GSE149683} \\ \midrule

Massively parallel scATAC (Satpathy et al., 2019) &
Annotation, Trajectory, Regulatory modeling &
Pioneering high-throughput scATAC-seq dataset of $\sim$200k immune and cancer nuclei enabling scalable mapping of open-chromatin landscapes; forms benchmark for lineage and immune-cell trajectory studies. &
$\sim$200k nuclei / blood and tumor tissues &
\href{https://www.ncbi.nlm.nih.gov/geo/query/acc.cgi?acc=GSE123581}{Nat.\ Biotechnol.\ 2019, GSE123581} \\ \midrule

ENCODE portal (single-cell experiments) &
Annotation, Integration, Regulatory modeling &
Centralized repository from the ENCODE Consortium aggregating thousands of single-cell RNA and ATAC assays from human and mouse tissues; provides uniformly processed metadata, peak calls, and cCRE annotations for benchmarking. &
$>$2,000 single-cell assays / multiple tissues &
\href{https://www.encodeproject.org/}{ENCODE Portal} \\ \midrule

T cell epigenetic atlas (Giles et al., 2022) &
Trajectory, GRN, Disease modeling &
Single-cell ATAC-seq atlas profiling chromatin accessibility across human T-cell activation, exhaustion, and differentiation states; supports trajectory reconstruction and immune-epigenetic modeling. &
$\sim$150k nuclei / T-cell subsets &
\href{https://www.nature.com/articles/s41590-022-01206-z}{Nat.\ Immunol.\ 2022} 

\end{longtable}
\endgroup

\vspace{1.5em} 
\begingroup

\setlength{\LTpre}{0pt}
\setlength{\LTpost}{0pt}

\begin{longtable}{@{}L{0.18\textwidth} L{0.16\textwidth} L{0.42\textwidth} L{0.12\textwidth} L{0.12\textwidth}@{}}
\caption{\textbf{Multiome} datasets (paired/tri-omic: RNA\,+\,ADT, RNA\,+\,ATAC, ATAC\,+\,ADT, and tri-modal) used in LLM-based single-cell research.}
\label{tab:datasets-multiome}\\
\toprule
\textbf{Dataset} & \textbf{Tasks} & \textbf{Description} & \textbf{Scale} & \textbf{Link / Citation} \\
\midrule
\endfirsthead
\toprule
\textbf{Dataset} & \textbf{Tasks} & \textbf{Description} & \textbf{Scale} & \textbf{Link / Citation} \\
\midrule
\endhead
\midrule \multicolumn{5}{r}{\emph{Continued on next page}} \\
\endfoot
\bottomrule
\endlastfoot

UCSC Cell Browser Hub &
Annotation, Integration, Visualization &
Aggregated repository of hundreds of public single-cell datasets across species and modalities with metadata and embeddings; useful for exploratory analysis and reference selection. &
$>$1{,}200 datasets / multi-species &
\href{https://cells.ucsc.edu/}{UCSC Cell Browser Hub} \\ \midrule

Human Cell Atlas (HCA) data browser (multi-project) &
Annotation, Integration, Cross-project mapping &
A global human cell atlas aggregating $\sim$63.2\,M cells across 515 projects and $>$11{,}000 donors, spanning diverse tissues and modalities; enables cross-study reference and meta-integration. &
$\sim$63.2\,M cells / 515 projects / 11{,}000$+$ donors &
\href{https://data.humancellatlas.org/}{HCA Data Portal} \\ \midrule

Azimuth reference collections (PBMC, lung, kidney, fetal) (RNA\,+\,ADT) &
Annotation, Integration &
Curated single-cell reference atlases by the Satija Lab for automated cell-type mapping in Seurat/Azimuth; harmonized labels and multimodal ADT features. &
100k–1\,M cells across multiple organs &
\href{https://azimuth.hubmapconsortium.org/}{Azimuth Data Portal} \\ \midrule

TEA-seq (tri-modal RNA\,+\,ATAC\,+\, ADT; GSE158013) &
Integration, Perturbation, Multi-modal reasoning &
Tri-modal PBMC dataset measuring transcriptome, chromatin accessibility, and surface proteins; supports alignment/fusion of omics layers and pretraining. &
$\sim$100k cells &
\href{https://www.ncbi.nlm.nih.gov/geo/query/acc.cgi?acc=GSE158013}{GSE158013} \\ \midrule

DOGMA-seq (RNA\,+\,ATAC\,+\,
Protein) &
Integration, Perturbation, Cross-modal reasoning &
Tri-modal profiling of transcriptome, chromatin, and proteins in human immune cells; benchmark for joint embeddings and cross-modal translation. &
$\sim$50k cells / PBMCs &
\href{https://www.ncbi.nlm.nih.gov/geo/query/acc.cgi?acc=GSE184715}{Nature Cell Biol.\ 2022 / GSE184715} \\ \midrule

ASAP-seq (ATAC\,+\,Protein) &
Integration, GRN, Epigenetic modeling &
Paired chromatin accessibility and surface proteins via antibody-derived tags; links cis-regulatory variation with immune phenotypes. &
$\sim$100k nuclei / PBMCs &
\href{https://www.ncbi.nlm.nih.gov/geo/query/acc.cgi?acc=GSE162690}{Nat.\ Biotechnol.\ 2021 / GSE162690} \\ \midrule

CITE-seq compendia (RNA\,+\,ADT) &
Annotation, Integration, Transfer learning &
Large compendium of paired RNA and surface-protein profiles across blood and tissue; enables multimodal foundation pretraining and zero-shot annotation. &
$>$500k cells across multiple tissues &
\href{https://www.ncbi.nlm.nih.gov/geo/query/acc.cgi?acc=GSE100866}{Nat.\ Methods 2019 / GSE100866} \\ \midrule

Multiome Benchmark Pack (QuKun Lab) &
Integration, Scalability, Batch-effect analysis &
Public benchmark suite consolidating 25 RNA\,+\,Protein, 12 RNA\,+\,ATAC, and 4 tri-omic datasets (CITE/TEA/DOGMA), standardized for cross-modal and large-scale integration testing. &
41 datasets total &
\href{https://github.com/QuKunLab/MultiomeBenchmarking}{Benchmark Portal} 

\end{longtable}
\endgroup

\vspace{1.5em} 
\begingroup

\setlength{\LTpre}{0pt}
\setlength{\LTpost}{0pt}

\begin{longtable}{@{}L{0.18\textwidth} L{0.16\textwidth} L{0.42\textwidth} L{0.12\textwidth} L{0.12\textwidth}@{}}
\caption{\textbf{Spatial transcriptomics and imaging} datasets used in LLM-based single-cell research.}
\label{tab:datasets-spatial}\\
\toprule
\textbf{Dataset} & \textbf{Tasks} & \textbf{Description} & \textbf{Scale} & \textbf{Link / Citation} \\
\midrule
\endfirsthead
\toprule
\textbf{Dataset} & \textbf{Tasks} & \textbf{Description} & \textbf{Scale} & \textbf{Link / Citation} \\
\midrule
\endhead
\midrule \multicolumn{5}{r}{\emph{Continued on next page}} \\
\endfoot
\bottomrule
\endlastfoot

Slide-seq / Slide-seqV2 &
Spatial mapping, Trajectory, Integration &
High-resolution spatial transcriptomics of mouse brain (hippocampus, cerebellum, cortex) at $\sim$10\,µm bead resolution; supports neighborhood and spatial-domain reconstruction benchmarks. &
$\sim$50k–100k spots per tissue &
\href{https://singlecell.broadinstitute.org/single_cell/study/SCP815}{Nat.\ Biotechnol.\ 2020 / SCP815} \\ \midrule

MERFISH (imaging) &
Spatial mapping, Pathway, Annotation &
Multiplexed error-robust fluorescence in situ hybridization (MERFISH) datasets profiling millions of mouse-brain cells with 3D coordinates and 483–1{,}000-gene panels; benchmark for high-resolution spatial mapping. &
$\sim$4\,M cells / 483–1{,}000 genes &
\href{https://vizgen.com/data}{Vizgen MERFISH Portal / Science 2018} \\ \midrule

Stereo-seq (STOmics) &
Spatial mapping, Developmental trajectory &
Genome-scale Stereo-seq datasets with submicron resolution; includes MOSTA (mouse embryo) and 3D \textit{Drosophila} atlases for developmental and cross-species modeling. &
$\sim$100\,µm $\times$ 100\,µm tiles / millions of spots &
\href{https://db.cngb.org/stomics/}{Cell 2022 / STOmics Data Hub} \\ \midrule

DBiT-seq \& spatial multi-omics &
Spatial mapping, Integration, Multi-omic reasoning &
Deterministic barcoding in tissue (DBiT-seq) capturing spatial RNA and protein expression in mouse embryo and human lymph node; supports multi-omic spatial benchmarks. &
$\sim$20k spots / 100\,µm grids &
\href{https://www.ncbi.nlm.nih.gov/geo/query/acc.cgi?acc=GSE152506}{Nat.\ Biotechnol.\ 2020 / GSE152506} \\ \midrule

10x Visium / Visium HD &
Spatial mapping, Annotation, Integration &
Widely used capture-array platform for spatial transcriptomics; Visium uses 55\,µm spots, Visium HD extends to 2\,µm grids with improved gene recovery and FFPE compatibility. &
$\sim$5k–55k spots per tissue &
\href{https://www.10xgenomics.com/spatial-transcriptomics}{10x Genomics Visium Portal} \\ \midrule

Xenium / CosMx (in situ) &
Spatial mapping, Clinical translation, Privacy &
In situ spatial transcriptomics platforms (10x Xenium, NanoString CosMx) profiling thousands of transcripts in human FFPE and fresh-frozen tissues (e.g., breast, colon, NSCLC); bridge omics and histology for clinical translation. &
10k–100k cells per section &
\href{https://www.10xgenomics.com/xenium}{Xenium Explorer} \,/\, \href{https://nanostring.com/products/cosmx-spatial-molecular-imager/}{CosMx Portal}

\end{longtable}
\endgroup

\vspace{1.5em} 
\begingroup

\setlength{\LTpre}{0pt}
\setlength{\LTpost}{0pt}

\begin{longtable}{@{}L{0.18\textwidth} L{0.16\textwidth} L{0.42\textwidth} L{0.12\textwidth} L{0.12\textwidth}@{}}
\caption{\textbf{Perturbation and Drug-Response} single-cell datasets used in LLM-based frameworks.}
\label{tab:datasets-perturb}\\
\toprule
\textbf{Dataset} & \textbf{Tasks} & \textbf{Description} & \textbf{Scale} & \textbf{Link / Citation} \\
\midrule
\endfirsthead
\toprule
\textbf{Dataset} & \textbf{Tasks} & \textbf{Description} & \textbf{Scale} & \textbf{Link / Citation} \\
\midrule
\endhead
\midrule \multicolumn{5}{r}{\emph{Continued on next page}} \\
\endfoot
\bottomrule
\endlastfoot

Genome-scale Perturb-seq (Replogle 2022) processed data &
Perturbation, GRN, Causal inference &
Largest CRISPR-based Perturb-seq dataset profiling $\sim$2.5 M single cells across $>$2 000 genetic perturbations; benchmark for causal network inference and representation learning. &
$\sim$2.5 M cells / $>$2 000 perturbations &
\href{https://plus.figshare.com/articles/dataset/_Mapping_information-rich_genotype-phenotype_landscapes_with_genome-scale_Perturb-seq_Replogle_et_al_2022_processed_Perturb-seq_datasets/20029387}{Cell 2022 / Figshare Dataset} \\ \midrule

Norman et al. 2019 Perturb-seq &
Perturbation, Combinatorial GRN &
Foundational combinatorial CRISPR Perturb-seq dataset (immune + cancer models); establishes feasibility of pooled functional genomics at single-cell resolution. &
$\sim$200 k cells / $>$250 combinations &
\href{https://www.ncbi.nlm.nih.gov/geo/query/acc.cgi?acc=GSE133344}{Science 2019 / GSE133344} \\ \midrule

Dixit et al. 2016 (GSE90063) &
Perturbation, Combinatorial GRN &
Early Perturb-seq study targeting 24 genes in macrophages; establishes pipeline for pooled CRISPR screens with single-cell RNA-seq. &
$\sim$100 k cells / 24 target genes &
\href{https://www.ncbi.nlm.nih.gov/geo/query/acc.cgi?acc=GSE90063}{Cell 2016 / GSE90063} \\ \midrule

Adamson et al. 2016 &
Perturbation, Stress-response, GRN &
Single-cell CRISPR Perturb-seq of ER-stress pathways in K562 cells; benchmark for pathway-level perturbation responses. &
$\sim$30 k cells / 10 perturbations &
\href{https://www.ncbi.nlm.nih.gov/geo/query/acc.cgi?acc=GSE90060}{Cell 2016 / GSE90060} \\ \midrule

sci-Plex collection (drug screens) + cellxgene portal &
Perturbation, Stress-response, GRN &
Multiplexed chemical-perturbation scRNA-seq of $>$650 k cells treated with 188 compounds via sci-Plex barcoding; pharmacotranscriptomic profiles for drug response modeling. &
$\sim$650 k cells / 188 drugs &
\href{https://cellxgene.cziscience.com/collections/3a41fbb2-7f3c-4ab2-8b5e-420a672f4d7f}{Cell 2020 / cellxgene Collection} \\ \midrule

Compressed Perturb-seq (immune LPS, 598 genes) &
Drug response, Perturbation, Benchmarking &
Low-multiplicity Perturb-seq library targeting 598 immune genes in macrophages under LPS stimulation; enables compressed experimental design for causal modeling. &
$\sim$300 k cells / 598 targets &
\href{https://www.ncbi.nlm.nih.gov/geo/query/acc.cgi?acc=GSE179924}{bioRxiv 2021 / GSE179924} \\ \midrule

In vivo Perturb-seq (brain / ASD genes) &
Perturbation, GRN, Neuroscience &
\textit{In vivo} CRISPR Perturb-seq targeting $\sim$30 ASD-linked genes in mouse cortex; maps neuronal gene-regulatory networks in native contexts. &
$\sim$200 k cells / 30 genes &
\href{https://www.ncbi.nlm.nih.gov/geo/query/acc.cgi?acc=GSE216113}{Science 2023 / GSE216113} \\ \midrule

Virtual Cell Challenge PBMC cytokine perturbations &
Perturbation, GRN, Benchmarking &
Community benchmark dataset (ARC Institute 2024) of $\sim$300 k PBMC cells under cytokine stimulation and CRISPRi perturbations; standardized splits for LLM and state-transition evaluation. &
$\sim$300 k cells / 150 perturbed genes &
\href{https://www.kaggle.com/competitions/virtual-cell-challenge}{ARC Challenge Repository / Kaggle} 

\end{longtable}
\endgroup

\vspace{1.5em} 
\begingroup

\setlength{\LTpre}{0pt}
\setlength{\LTpost}{0pt}

\begin{longtable}{@{}L{0.18\textwidth} L{0.16\textwidth} L{0.42\textwidth} L{0.12\textwidth} L{0.12\textwidth}@{}}
\caption{\textbf{Plant RNA} single-cell transcriptomics datasets used in LLM research.}
\label{tab:datasets-plant}\\
\toprule
\textbf{Dataset} & \textbf{Tasks} & \textbf{Description} & \textbf{Scale} & \textbf{Link / Citation} \\
\midrule
\endfirsthead
\toprule
\textbf{Dataset} & \textbf{Tasks} & \textbf{Description} & \textbf{Scale} & \textbf{Link / Citation} \\
\midrule
\endhead
\midrule \multicolumn{5}{r}{\emph{Continued on next page}} \\
\endfoot
\bottomrule
\endlastfoot

scPlantDB (meta-collection) &
Annotation, Integration, Meta-analysis &
Comprehensive plant single-cell transcriptome database integrating 67 datasets from 17 plant species ($\sim$2.5 M cells); unified preprocessing and annotations; supports cross-species modeling and plant-specific LLMs. &
$\sim$2.5 M cells / 67 datasets / 17 species &
\href{https://biobigdata.nju.edu.cn/scplantdb/home}{scPlantDB Portal} \\ \midrule

PlantscRNAdb &
Annotation, Marker discovery &
Curated database of plant cell-type marker genes across four species (Arabidopsis, rice, maize, tomato); supports ontology-based annotation and cell identity benchmarking. &
4 species / multiple tissues &
\href{http://ibi.zju.edu.cn/plantscrnadb/}{PlantscRNAdb Portal} \\ \midrule

Arabidopsis scRNA-seq (E-CURD-4) &
Annotation, Trajectory, Development &
Baseline scRNA-seq dataset of Arabidopsis thaliana root and leaf tissues ($\sim$10,779 cells); used for developmental lineage and differentiation analysis. &
10,779 cells / 2 tissues &
\href{https://www.ebi.ac.uk/gxa/sc/experiments/E-CURD-4/results/cell-plots}{EBI ArrayExpress E-CURD-4} \\ \midrule

Tobacco leaf scRNA-seq (transgenic antibody line) &
Annotation, Perturbation, Stress response &
Single-cell transcriptomic profiling of transgenic \textit{Nicotiana tabacum} leaves expressing llama antibody; identifies immune-like responses and cell-type heterogeneity under genetic perturbation. &
$\sim$25k cells / leaf tissue &
\href{https://www.nature.com/articles/s41597-023-02833-5}{Sci.\ Data 2023} \\ \midrule

Plant scRNA Browser (PscB) &
Annotation, Visualization, Cross-tissue integration &
Online visualization hub aggregating plant scRNA-seq datasets (Arabidopsis, rice, \textit{Wolffia}) with harmonized metadata and interactive UMAP-based cell-type search. &
15+ datasets / 3+ species &
\href{https://www.zmbp-resources.uni-tuebingen.de/timmermans/plant-single-cell-browser/}{PscB Data Portal} 

\end{longtable}
\endgroup

\section{Methods Comparison}

\begingroup
\setlength{\LTpre}{0pt}
\setlength{\LTpost}{0pt}
\setlength{\tabcolsep}{3.5pt}   
\renewcommand{\arraystretch}{1.07}
\sloppy                         

\begin{longtable}{@{}
  L{0.15\textwidth}  
  L{0.09\textwidth}  
  L{0.11\textwidth}  
  L{0.12\textwidth}  
  L{0.12\textwidth}  
  C{0.07\textwidth}  
  L{0.25\textwidth}  
  C{0.06\textwidth}  
@{}}
\caption{Full comparison of single-cell LLM and agentic methods (Domains column removed). Year merged into Model.}\label{tab:methods-full}\\
\toprule
\textbf{Model (Year)} & \textbf{Published where} & \textbf{Category} & \textbf{Modality} & \textbf{Grounding Type} & \textbf{Agentic (Y/N)} & \textbf{Primary Task} & \textbf{Domain Score} \\
\midrule
\endfirsthead
\toprule
\textbf{Model (Year)} & \textbf{Published where} & \textbf{Category} & \textbf{Modality} & \textbf{Grounding Type} & \textbf{Agentic (Y/N)} & \textbf{Primary Task} & \textbf{Domain Score} \\
\midrule
\endhead
\midrule \multicolumn{8}{r}{\emph{Continued on next page}} \\
\endfoot
\bottomrule
\endlastfoot
scGPT \cite{cui2024scgpt} & Nature Methods & Foundation & Multiomics (scRNA + optionally multi-omic modes) & Atlas (trained on large atlas of single-cell datasets) & No & Annotation, Integration, Perturbation (annotation as primary) & 6 \\ \midrule
Geneformer \cite{theodoris2023transfer} & Nature & Foundation & scRNA & None / Rank-based (implicit) & No & Cell classification, in-silico perturbation, network prediction & 6 \\ \midrule
scFoundation \cite{hao2024large} & Nature Methods  & Foundation & scRNA / multi-omics (transcriptomics focus) & Atlas (large pretrained corpus) & No & Cell annotation, perturbation prediction, drug response, gene module inference & 7 \\ \midrule
CellFM \cite{zeng2025cellfm} & Nature Communications & Foundation & scRNA & Atlas / value-projection & No & Learns embeddings from 100M cells and supports annotation, perturbation, gene function & 7 \\ \midrule
iSEEEK \cite{shen2022universal} & Briefings in Bioinformatics & Foundation & scRNA & Atlas (11.9M cells, human + mouse) & No & Integrates massive single-cell datasets via gene-ranking similarity to enable scalable cross-dataset embedding and clustering & 5 \\ \midrule
tGPT \cite{shen2023generative} & iScience & Foundation Model (Autoregressive) & scRNA & Atlas (22.3 M cells) & No & Treats gene-expression ranks as token sequences and learns generative embeddings for cell clustering, trajectory inference, and bulk-tissue analysis & 5 \\ \midrule
scBERT \cite{yang2022scbert} & Nature Machine Intelligence & Foundation Model (Encoder-only) & scRNA & Atlas (public scRNA-seq corpora) & No & Learns bidirectional gene–cell embeddings using BERT-style masked modeling for robust cell-type annotation and novel cell discovery & 4 \\ \midrule
UCE (Universal Cell Embeddings) \cite{rosen2023universal} & BioRxiv & Foundation & scRNA (cross-species) & Atlas / binary masked self-supervision & No & Embeds any cell zero-shot across species into a shared latent space for clustering, lineage inference, annotation & 6 \\ \midrule
GeneCompass \cite{yang2024genecompass} & Nature-/Cell Research & Foundation Model & scRNA / cross-species & Ontology (GRN + co-expression + gene family knowledge) & No & Learns cross-species gene–cell embeddings and supports annotation, perturbation, dose-response, and GRN inference & 6 \\ \midrule
scELMo \cite{liu2023scelmo} & BioRxiv & Text-Bridge LLM & scRNA + Metadata Text & LLM-generated embeddings from gene/metadata descriptions + raw data & No & Embeds cells via text-derived gene metadata embeddings combined with expression, then supports clustering, batch correction, annotation, perturbation & 5 \\ \midrule
CellPLM \cite{wen2023cellplm} & ICLR & Spatial (Multimodal Foundation Model) & scRNA + Spatial (SRT) & Atlas + Spatial relations (uses SRT in pretraining) & No & Treats cells as tokens and tissues as sentences, leveraging spatially-resolved transcriptomics and a Gaussian-mixture prior to encode inter-cell relations for denoising, spatial imputation, and perturbation prediction.  & 4 \\ \midrule
scMoFormer \cite{tang2023single} & ArXiv & Spatial (Multimodal Foundation Model) & scRNA + Protein / Multi-omics & Atlas + domain knowledge in cross-modality aggregation & No & Uses modality-specific transformers and cross-attention to impute missing modalities, classify cells, and fuse multimodal representations & 4 \\ \midrule
scFormer \cite{xu2024scmformer} & ArXiv & Spatial (Multimodal Foundation Model) (or Foundation + multi-omics) & Transcript-
omics + Proteomics / multimodal & Atlas + transformer-based fusion & No & Aligns and integrates multi-omics single-cell data, recovers missing modalities, and transfers labels across modalities & 4 \\ \midrule
scMulan \cite{bian2024scmulan} & BioRxiv & Foundation Model & scRNA + Metadata & Atlas + prompt-conditioned generative modeling & No & Encodes each cell as a “c-sentence” integrating expression + metadata; supports zero-shot annotation, batch integration, and conditional generation & 3 \\ \midrule
scPRINT \cite{kalfon2025scprint} & Nature Communications & Foundation Model & scRNA & Atlas (50M+ cell pretraining) & No & Learns cell embeddings and infers cell-specific gene regulatory networks; supports zero-shot denoising, batch correction, label prediction, expression reconstruction & 5 \\ \midrule
scGraphformer \cite{fan2024scgraphformer} & Nature Communications Biology & Foundation Model & scRNA & Atlas + relational prior (kNN bias, refined) & No & Learns a cell–cell graph via a transformer-GNN hybrid for better classification and interaction inference & 4 \\ \midrule
scRDiT \cite{dong2025scrdit} & ArXiv & Foundation Model & scRNA (transcriptome) & Atlas-like generative prior / diffusion modeling & No & Generates synthetic scRNA-seq samples via diffusion transformer + DDIM for accelerated sampling & 2 \\ \midrule
scGFT \cite{nouri2025single} & Nature Communications Biology & Foundation Model & scRNA & Fourier-based perturbation / reconstruction (train-free) & No & Synthesizes new single-cell expression profiles by perturbing Fourier components in frequency space. & 4 \\ \midrule
scTrans \cite{lu2024sctrans} & IJCAI & Foundation Model & scRNA & Sub-vector masked completion over gene modules & No & Learns multi-scale sub-vector tokens to perform gene-selective cell-type annotation via masked completion and contrastive regularization & 3 \\ \midrule
scGT  (Graph Transformer) \cite{qi2025scgt} & Bioinfo-rmatics Advance & Graph / Multi-omics Integration & scRNA + scATAC & Observed / Hybrid Graph & No & Multi-omics integration + label transfer & 4 \\ \midrule
TransformerST \cite{lu2024sctrans} & Briefings in Bioinformatics & Spatial / Multimodal FM & Spatial (histology + gene expression) & Spatial (histology + gene expression) & No & Super-resolution gene expression \& tissue clustering & 3 \\ \midrule
scGPT-spatial \cite{cui2024scgpt} & BioRxiv & Spatial (Multimodal Foundation Model) & Spatial Transcriptomics + scRNA prior & Atlas + spatial-aware decoding & No & Extends scGPT via continual pretraining to spatial data, supports multi-slide integration, cell-type deconvolution, and spatial gene imputation & 3 \\ \midrule
HEIST \cite{madhu2025heist} & BioRxiv & Spatial (Multimodal Foundation Model) & Spatial transcriptomics + proteomics & Hierarchical graph modeling (spatial + GRNs) & No & Learns joint embeddings of cells and genes in spatial context to perform cell annotation, gene imputation, spatial clustering, clinical outcome prediction & 4 \\ \midrule
stFormer \cite{cao2024stformer} & BioRxiv & Spatial (Multimodal Foundation Model) & Spatial transcriptomics + ligand context & Atlas + biased cross-attention to ligand niche genes & No & Learns gene embeddings contextualized by ligand signals in spatial microenvironments; aids clustering, ligand–receptor inference, and perturbation simulation & 4 \\ \midrule
FmH2ST \cite{wang2025fmh2st} & Nucleic Acids Research & Spatial (Multimodal Foundation Model) & Histology image + spatial transcriptomics & Image foundation + dual graphs + spot branch & No & Predict spatial gene expression from histology using fused image and spot features, supporting denoising, heterogeneity detection, and regulatory inference & 3 \\ \midrule
OmiCLIP \cite{cui2025towards} & Nature Methods & Spatial (Multimodal Foundation Model) & Histology + Spatial Transcriptomics & Image–gene contrastive alignment (rank-based) & No & Learns unified visual-omics embeddings linking histopathology and spatial gene expression; enables cross-modal prediction, annotation, and tissue retrieval & 7 \\ \midrule
QuST-LLM \cite{huang2024qust} & ArXiv & Text-Bridge LLM & Spatial transcriptomics + histology metadata & GO-term + gene enrichment + LLM narrative overlay & Yes & Converts ST data and ROIs into human-readable narratives and matches natural language queries to spatial regions via GO/LLM interpretation & 4 \\ \midrule
SpaCCC \cite{yang2024deciphering} & IEEE Xplore & Text-Bridge LLM & Spatial transcriptomics / transcriptomic genes (LRs) & LLM embeddings of ligand + receptor genes & No & Infers spatially resolved cell–cell communication by embedding LR pairs in LLM latent space + diffusion / permutation test filtering & 3 \\ \midrule
spaLLM \cite{10778152} & Briefings in Bioinformatics & Spatial (Multimodal Foundation Model) & Spatial multi-omics (RNA, ATAC, protein) & scGPT embeddings + GNN + attention fusion & No & Enhances spatial domain identification by fusing LLM-derived embeddings and spatial-omics signals via multi-view attention & 3 \\ \midrule
scMMGPT (2025) \cite{shi2025multimodal} & ArXiv & Text-Bridge LLM / Multimodal\_FM & RNA + Text (metadata) & Annotation, description & No & Generation - links single-cell and text PLMs to describe cells, generate pseudo-cells from text, and enhance annotation through text-conditioned reasoning & 6 \\ \midrule
Cell2Sentence (C2S) \cite{pmlr-v235-levine24a} & PMLR & Text-Bridge LLM & RNA (scRNA-seq converted to “cell sentences”) & Marker / Implicit (gene rank-order) & No & Generation, annotation \& reconstruction - encodes cells as gene-ranked sentences and fine-tunes LLMs to classify or generate biologically meaningful cell text & 6 \\ \midrule
C2S-Scale \cite{Rizvi2025.04.14.648850} & BioRxiv & Text-Bridge LLM+ Multimodal Foundation Model & RNA + Text / Metadata & Atlas + Text / Implicit rank grounding & No & Chat, generation \& annotation - trains large LLMs on cell sentences and biological text to enable perturbation prediction and multicellular summarization & 7 \\ \midrule
Cell2Text \cite{kharouiche2025cell2text} & ArXiv  & Text-Bridge LLM + Multimodal Foundation Model & RNA → Natural Language & Gene-level embeddings + ontology / metadata grounding & No & Expression prediction \& regulatory inference - learns regulatory syntax from chromatin accessibility and DNA motifs to predict expression and interpret TF–cis interactions & 7 \\ \midrule
GenePT \cite{chen2024genept} & BioRxiv & Foundation Model / Text-Augmented & scRNA & Literature / Text embedding grounding & No & Embedding \& downstream prediction - uses GPT-3.5 gene text embeddings to derive cell embeddings via weighted or ranked gene aggregation. & 5-7 \\ \midrule
CellLM \cite{zhao2023large} & ArXiv & Foundation Model / Representation model & RNA (single cell expression) & Implicit - contrastive embedding from expression data & No & Represent cells via a contrastive-learning transformer, optimize embedding space for tasks like cell-type annotation, drug sensitivity prediction & 6 \\ \midrule
scExtract \cite{wu2025scextract} & Genome Biology & Agentic Framework (Text-Bridge LLM) & RNA (scRNA-seq) + Text & Article-based parameter extraction & Yes & Annotation \& Integration - uses LLMs to extract pipeline parameters from publications and apply them for dataset harmonization. & 7 \\ \midrule
scAgent \cite{mao2025scagent} & ArXiv  & Agentic Framework  (Text-Bridge hybrid) & RNA (scRNA-seq) & Reference atlas + marker gene reasoning + memory grounding & Yes & Universal cell annotation \& novel cell discovery - scAgent uses an LLM planning module, memory, and tool modules to annotate cells across tissues, detect unknown types, and incrementally learn new annotations & 7 \\ \midrule
EpiFoundation \cite{wu2025epifoundation} & BioRxiv & Epigenomic Foundation Model / Multimodal alignment & scATAC (chromatin accessibility) & Peak-to-gene supervision (alignment to expression) & No & Cell embedding, annotation, batch correction, gene expression prediction - trains on sparse peak sets, aligns to gene expression supervision, and transfers learned embeddings for downstream ATAC tasks & 7 \\ \midrule
EpiAgent \cite{chen2025epiagent} & BioRxiv & Epigenomic Foundation Model / Multimodal text-bridge hybrid & scATAC / chromatin accessibility & Peak tokenization + external embeddings + regulatory supervision grounding & No & Embedding, annotation, imputation, and perturbation prediction - encodes chromatin accessibility as ranked cCRE tokens, enabling zero-shot cell annotation, peak imputation, and response prediction. & 8 \\ \midrule
ChromFound \cite{jiao2025chromfound} & Nature & Epigenomic → Expression Transformer / Regulatory FM & scATAC (chromatin accessibility) + DNA sequence & Motif × peak matrix / masked regulatory grammar grounding & No & Expression prediction \& regulatory inference - learns regulatory syntax from chromatin accessibility and DNA motifs to predict gene expression in seen and unseen cell types; also interprets transcription factor interactions and cis-regulatory elements & 8 \\ \midrule
GET (General Expression Transformer) \cite{fu2025foundation} & Nature & Text-Bridge LLM (Hybrid Fusion Model) & RNA (scRNA) / Text embeddings & Fusion of scGPT embeddings + text-encoded “cell sentences” grounding & No & Annotation (fusion embedding model) - combines scGPT-derived embeddings and LLM (text encoder) embeddings via a small fusion MLP to improve cell-type classification robustness across datasets & 6 \\ \midrule
scMPT \cite{palayew2025towards} & ArXiv & Text-Bridge LLM (Hybrid Fusion Model) & RNA (scRNA) / Text embeddings & Fusion of scGPT embeddings + text-encoded “cell sentences” grounding & No & Annotation (fusion embedding model) - combines scGPT-derived embeddings and LLM (text encoder) embeddings via a small fusion MLP to improve cell-type classification robustness across datasets & 6 \\ \midrule
EpiBERT \cite{javed2025multi} & Cell Genomics & Epigenomic Foundation Model / Multi-modal Transformer & DNA sequence + chromatin accessibility & Masked-accessibility pretraining + motif \& sequence fusion & No & Accessibility imputation, gene expression prediction \& regulatory inference - predicts masked ATAC signals, then fine-tunes to predict expression and enhancer-gene links, generalizing to unseen cell types & 8 \\ \midrule
scMamba \cite{yuan2025scmamba} & ArXiv & Multimodal / Foundation\_FM & Multi-omics (RNA + others) & Implicit via integrated features; no explicit external grounding & No & Multi-omic embedding, integration \& annotation across modalities without prior feature selection & 7 \\ \midrule
GeneMamba \cite{qi2025bidirectional} & ArXiv & Foundation Model / State-Space Model & RNA (scRNA) & Implicit (via gene sequence context + pathway loss) & No & Multi-batch integration, cell-type annotation, gene correlation - uses a BiMamba state-space architecture with pathway-aware losses for scalable, context-rich modeling of scRNA & 7 \\ \midrule
Nicheformer \cite{schaar2024nicheformer} & BioRxiv & Foundation \_FM/ Spatial / Multimodal\_FM & scRNA + spatial transcriptomics & Contextual tokenization + metadata + spatial neighborhood embedding & No & Spatial context prediction, spatial label / niche prediction, mapping spatial info to dissociated cells & 7 \\ \midrule
scFormer Cell+ \cite{cui2022scformer} & Bioarxiv & Foundation Model / Multimodal Transformer & RNA (+ metadata) & Joint gene-cell embedding with metadata tokens & No & Integration \& Annotation - context-aware joint embedding for cross-species/tissue generalization & 7 \\ \midrule
scPlantLLM \cite{scplantlmm} & Genomics, Proteomics and Bioinformatics & Text-Bridge LLM (Foundation Model) & RNA (plant scRNA-seq) & Gene token + binned expression embedding (Plant Atlas grounding) & No & Annotation, Integration, GRN inference - pretrains on plant single-cell data with masked LM and cell-type supervision, enabling cross-species annotation, clustering, and regulatory discovery in plant systems & 7 \\ \midrule
LICT \cite{ye2024objectively} & ArXiv & Text-Bridge LLM / Annotation LLM Hybrid & RNA (scRNA-seq) & Marker gene–based DE lists + LLM prompting & No & Annotation \& Reliability - iterative LLM prompting with DE markers for label refinement and confidence scoring & 5 \\ \midrule
Teddy (family of models) \cite{chevalier2025teddy} & ArXiv & Foundation Model / Disease-aware Transformer & scRNA (single-cell RNA-seq) & Self-supervised + supervised annotation supervision & No & Disease state classification / healthy vs diseased detection - trained to identify disease conditions of held-out donors and distinguish diseased vs healthy cells in new disease contexts & 7 \\ \midrule
Pilot \cite{joodaki2024detection} & Github & Benchmark / Evaluation Framework & RNA & Model-agnostic pilot foundation testing & No & Benchmarking \& Evaluation - lightweight pilot framework for early single-cell FM testing & 4 \\ \midrule
CellVerse \cite{zhang2025cellverse} & ArXiv & Benchmark / Evaluation & Multi-omics (RNA, CITE, ASAP, etc.) & Implicit via prompt encoding & No & QA benchmark for annotation, drug response, perturbation tasks & 5 \\ \midrule
xTrimoGene \cite{gong2023xtrimogene} & NeurIPS / ArXiv & Foundation Model (Scalable Transformer) & scRNA-seq & Sparse masking + auto-discretization & No & Representation learning + annotation, perturbation, drug synergy prediction & 7 \\ \midrule
EpiAttend \cite{li2022epiattend} &  NeurIPS 2022 Workshop & Regulatory / Sequence-Epigenome Transformer & DNA sequence + single-cell epigenomic data & Sequence + cell-specific epigenome grounding & No & Predict cell type–specific gene expression by integrating DNA sequence and single-cell epigenomic tracks, linking enhancers and promoters & 6 \\ \midrule

Spatial2-Sentence \cite{chen2025spatial} & ArXiv & Spatial / Text-Bridge hybrid & Spatial + expression (Imaging Mass Cytometry) & Spatial adjacency + expression similarity tokenization & No & Encode spatial \& expression context into multi-sentence prompts for LLMs to perform cell-type classification and clinical status prediction & 6 \\ \midrule
ChatCell \cite{fang2024chatcell} & ArXiv & Text-Bridge LLM (Instructional) & scRNA → “cell sentence” & Vocabulary adaptation + unified sequence generation of cell sentences & No & Natural language interface for single-cell tasks - allows users to query, annotate, generate, and explore scRNA data via text prompts. Hugging Face & 6 \\ \midrule
CellAtria \cite{nouri2025agentic} & BioRxiv & Agentic Framework & scRNA- seq + metadata & Ontology (graph + metadata)& Yes & Annotation \& Ontology Mapping & 7 \\ \midrule

CellAgent \cite{xiao2024cellagent} & ArXiv & Agentic Framework & scRNA- seq & None & Yes & Annotation \& Ontology Mapping & 7

\end{longtable}
\endgroup

\clearpage
\section{Appendix Figure}
\begin{figure*}[t]
    \centering
    \includegraphics[width=\textwidth]{figure/model_task_heatmap_darkblueyellow.png}
    \caption{
        Task vs Model Heatmap
    }
    \label{fig:heatmap}
\end{figure*}

\begin{figure}[t]
    \centering
    \includegraphics[width=0.95\linewidth]{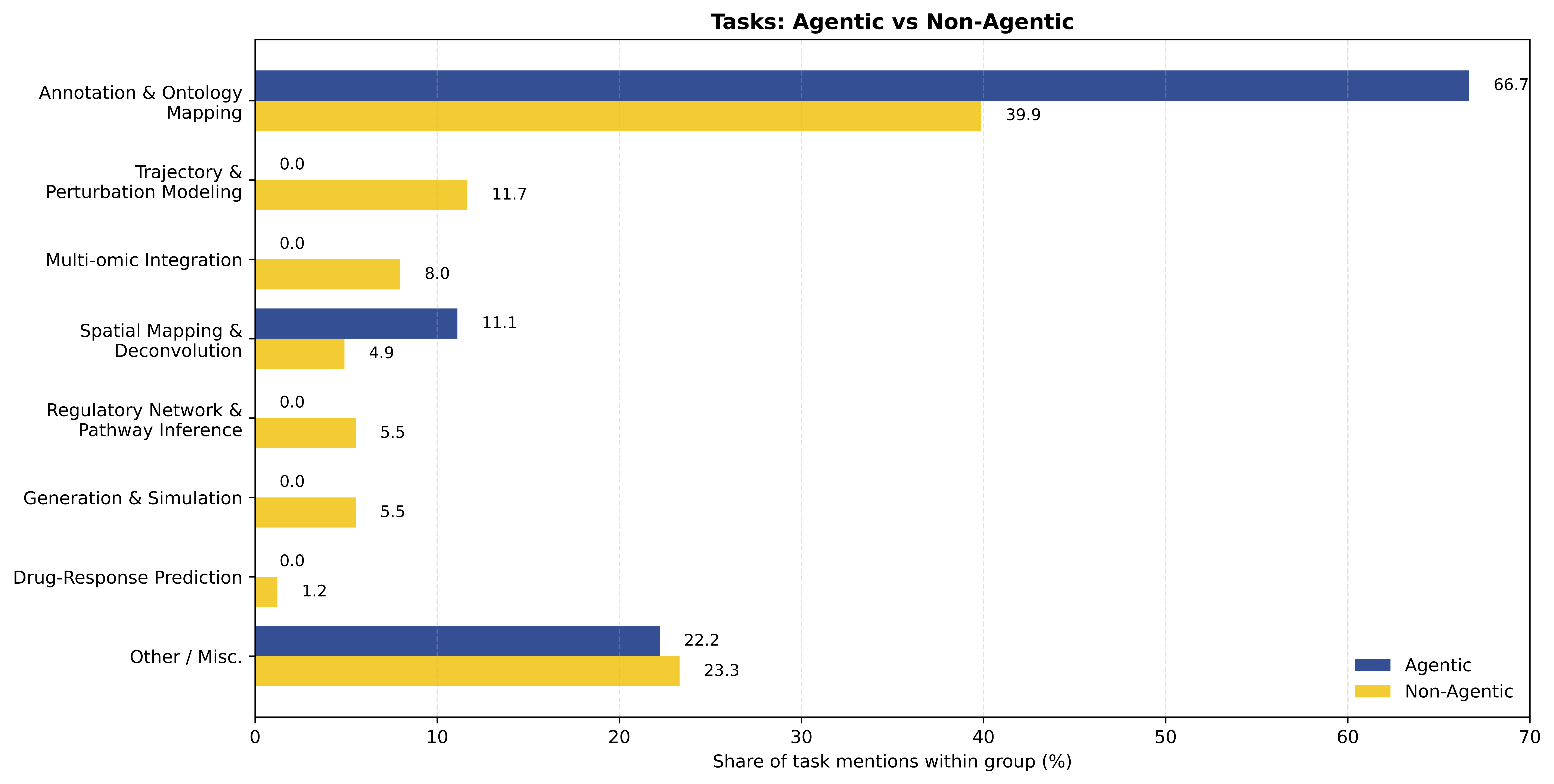}
    \caption{Comparison of task coverage between \textbf{agentic} (dark blue) and \textbf{non-agentic} (yellow) models. 
    Agentic frameworks emphasize annotation, ontology mapping, spatial mapping while non-agentic models concentrate on trajectory, perturbation modeling, regulatory and pathway inference as well.}
    \label{fig:agentic_tasks}
\end{figure}

\begin{figure}[t]
    \centering
    \includegraphics[width=0.95\linewidth]{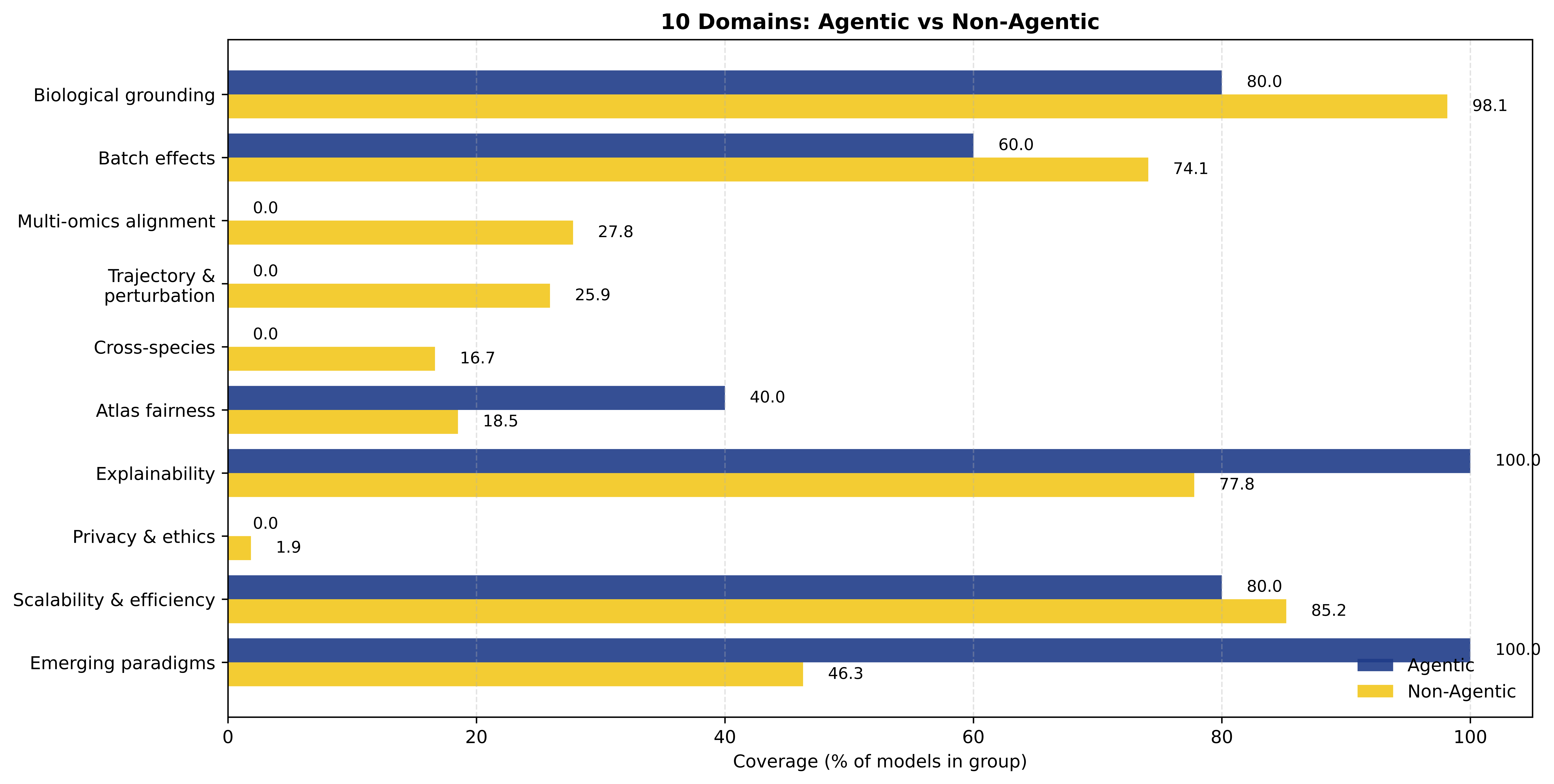}
    \caption{Comparison of domain coverage between \textbf{agentic} (dark blue) and \textbf{non-agentic} (yellow) models. 
    Agentic frameworks emphasize explainability, fairness, and emerging paradigms, 
    while non-agentic models concentrate on biological grounding and batch effects.}
    \label{fig:agentic_domains}
\end{figure}

\end{document}